\documentclass{article}

\usepackage[final, nonatbib]{neurips_2025}

\usepackage[utf8]{inputenc} 
\usepackage[T1]{fontenc}    
\usepackage[hidelinks]{hyperref}       
\usepackage{url}            
\usepackage{booktabs}       
\usepackage{amsfonts}       
\usepackage{nicefrac}       
\usepackage{microtype}      
\usepackage{xcolor}         
\usepackage[numbers]{natbib}

\usepackage{amsmath,amsfonts,bm}

\def\eqref#1{equation~\ref{#1}}

\def\1{\bm{1}}

\DeclareMathAlphabet{\mathsfit}{\encodingdefault}{\sfdefault}{m}{sl}
\SetMathAlphabet{\mathsfit}{bold}{\encodingdefault}{\sfdefault}{bx}{n}

\usepackage{comment}
\usepackage{xspace}
\usepackage{xcolor}
\usepackage{booktabs}

\usepackage{graphicx}
\usepackage[center]{subfigure}
\usepackage{balance}  

\usepackage{graphics}
\usepackage{xspace}

\usepackage{booktabs}
\usepackage{soul}
\usepackage{balance}

\usepackage{float}
\usepackage{algorithm}
\usepackage[noend]{algpseudocode}
\usepackage{amsmath}
\usepackage{amssymb}
\usepackage{mathtools}
\usepackage{amsfonts}
\usepackage{amsbsy}
\usepackage{amsthm}
\usepackage[]{caption}
\usepackage{multirow}
\usepackage[mathscr]{eucal} 
\usepackage{mathtools}
\usepackage{verbatim}
\usepackage{tabularx}
\usepackage{enumitem}
\usepackage{array}
\usepackage{soul}
\usepackage{dsfont}
\usepackage{url}
\usepackage{bbm}
\usepackage{wrapfig}
\usepackage{varwidth}

\usepackage[most]{tcolorbox}

\usepackage{pifont}
\usepackage{xcolor,colortbl}
\usepackage[inkscapeformat=png]{svg}
\usepackage{makecell}
\usepackage{xspace}
\usepackage{footmisc} 
\usepackage{wrapfig}
\hypersetup{
    colorlinks=true,
    linkcolor=black,
    citecolor = black,
    filecolor=magenta,      
    urlcolor=magenta,
    pdftitle={Overleaf Example},
    pdfpagemode=FullScreen,
    }

\definecolor{googleblue}{RGB}{66,133,244}
\definecolor{googlered}{RGB}{219,68,55}
\definecolor{googleyellow}{RGB}{244,180,0}
\definecolor{googlegreen}{RGB}{15,157,88}

\newtcolorbox{mybox}[2][]{%
  attach boxed title to top left
               = {yshift=-8pt},
  colback      = cyan!5!white,
  colframe     = googleblue,
  fonttitle    = \bfseries,
  colbacktitle = googleblue,
  title        = #2,#1,
  enhanced,
}

\newcommand{\method}{\textsc{FGP}\xspace}
\usepackage{xspace} 

\newcommand{\std}{\scriptsize}

\makeatletter
\newtheorem*{rep@theorem}{\rep@title}
\newcommand{\newreptheorem}[2]{%
\newenvironment{rep#1}[1]{%
 \def\rep@title{#2 \ref{##1}}%
 \begin{rep@theorem}}%
 {\end{rep@theorem}}}
\makeatother

\newreptheorem{theorem}{Theorem}

\definecolor{sunwoogreen}{rgb}{0.8, 0.8, 1.0}
\definecolor{sunwoogreen2}{RGB}{67, 148, 58}
\definecolor{sunwooyellow}{rgb}{1.0, 1.0, 0.0}
\definecolor{sunwooyellow2}{RGB}{228, 208, 10}
\definecolor{crimson}{RGB}{220, 20, 60}

\newcommand{\best}{\cellcolor{sunwoogreen}}  

\AtBeginDocument{%
  \providecommand\BibTeX{{%
    \normalfont B\kern-0.5em{\scshape i\kern-0.25em b}\kern-0.8em\TeX}}}

\title{Learning to Flow from Generative Pretext Tasks for Neural Architecture Encoding}

\author{
  Sunwoo Kim$^{1}$, Hyunjin Hwang$^{1}$, Kijung Shin$^{1,2}$ \\
  $^{1}$Kim Jaechul Graduate School of AI, $^{2}$School of Electrical Engineering \\
  Korea Advanced Institute of Science and Technology (KAIST)\\
  \texttt{\{kswoo97, julia510, kijungs\} @ kaist.ac.kr} \\
}

\begin{document}

\maketitle

\begin{abstract}
  The performance of a deep learning model on a specific task and dataset depends heavily on its neural architecture, motivating considerable efforts to rapidly and accurately identify architectures suited to the target task and dataset.
To achieve this, researchers use machine learning models—typically neural architecture encoders—to predict the performance of a neural architecture. 
Many state-of-the-art encoders aim to capture information flow within a neural architecture, which reflects how information moves through the forward pass and backpropagation, via a specialized model structure. 
However, due to their complicated structures, these flow-based encoders are significantly slower to process neural architectures compared to simpler encoders, presenting a notable practical challenge.
To address this, we propose FGP, a novel pre-training method for neural architecture encoding that trains an encoder to capture the information flow without requiring specialized model structures.
FGP trains an encoder to reconstruct a flow surrogate, our proposed representation of the neural architecture's information flow.
Our experiments show that FGP boosts encoder performance by up to 106\% in Precision@1\%, compared to the same encoder trained solely with supervised learning.
\end{abstract}

\section{Introduction}
\label{sec:introduction}

Deep learning has achieved outstanding performance across diverse machine learning tasks, including those in computer vision~\citep{dosovitskiy2020image, zhu2024vision}. 
However, a particular neural architecture (i.e., deep-learning model) that performs well on one task or dataset may underperform on others, highlighting the importance of choosing an architecture suited to the target task and dataset~\citep{liu2019darts, elsken2019neural, ou2024towards}.
A naive approach—exhaustively training and evaluating each neural architecture—incurs substantial costs since each architecture requires expensive computational resources and time for training and evaluation on the target task and dataset.

A promising approach for alleviating expensive training and evaluation costs is to use machine learning techniques to \textit{predict} the performance of neural architectures.
Due to their effectiveness, performance predictors are widely leveraged for neural architecture search to find good neural architectures rapidly and accurately~\citep{dudziak2020brp, white2021powerful, jawahar2023llm, ning2021evaluating}.
Recently, various neural architecture encoders, which are typically deep learning models, have been developed to be used for performance predictors~\cite{ning2022ta, yi2023nar, hwang2024flowerformer, ning2020generic}. 

The focus of many state-of-the-art neural architecture encoders~\citep{ning2022ta, hwang2024flowerformer} lies in capturing the \textit{information flow} within a neural architecture—a concept that describes how input data is propagated in the forward pass and how gradients flow during the backpropagation.
To capture the information flow, these flow-based encoders use complex model structures, especially graph neural networks with asynchronous message passing~\citep{thost2021directed}, specially designed to capture the flow within an architecture.




\begin{figure*}[t]
  \centering
  \subfigure[\small The total processing time of FlowerFormer~\citep{hwang2024flowerformer} and ResGatedGCN~\citep{bresson2017residual} for the 14K and 15K architectures in Nas-Bench-10 and NAS-Bench-201, respectively.]{%
    \includegraphics[width=0.5\textwidth]{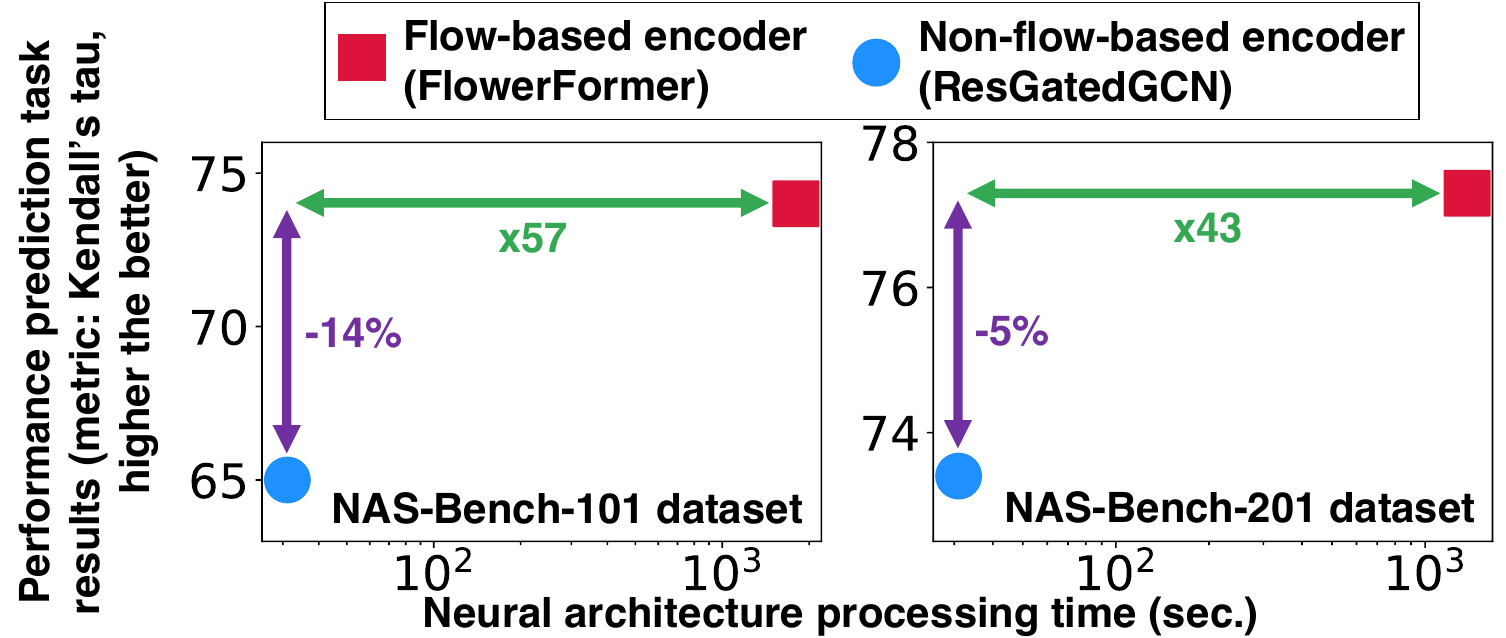}
    \label{fig:flowawarechallenge}
  }
  \hspace{1mm}
  \subfigure[\small Example generative pretext tasks: deep learning models are trained to predict the masked atoms or operations (colored in gray).]{%
    \includegraphics[width=0.4\textwidth]{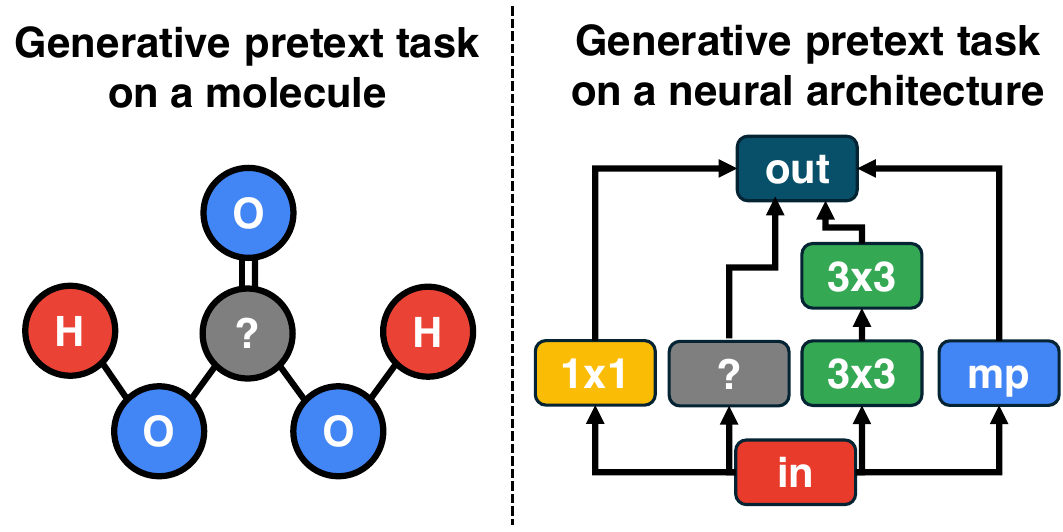}
    \label{fig:taskobjective}
  }
  \caption{\textbf{Remaining challenges in neural architecture encoding.}
  \textbf{Challenge 1}: Regarding model structures, as shown in (a), flow-based encoders, though more effective, are considerably slower than non-flow-based encoders.
  \textbf{Challenge 2}: Regarding pretext objectives, in (b), while a deep learning model learns chemical rules by identifying masked atoms in a molecule, it gains no clear training guidance when identifying masked operations in a neural architecture, where most operations are possible options due to the absence of explicit rules.}
  \label{fig:firstmotivation}
\end{figure*}

\noindent\textbf{Challenge 1.} 
However, these flow-based encoders face a practical challenge related to efficiency.
The complicated structures of flow-based encoders result in significantly longer processing times compared to the simpler structures of non-flow-based encoders. 
In particular, flow-based encoders \textit{sequentially} perform message passing according to the graph’s topological order (see Section~\ref{subsec:orderassignment}), instead of computing all messages simultaneously as in non-flow-based encoders; this sequential process results in significantly greater computation time.
Specifically, as shown in Figure~\ref{fig:firstmotivation} (a), a flow-based encoder (spec., FlowerFormer \citep{hwang2024flowerformer}) is up to 57 times slower than non-flow-based encoders~\citep{bresson2017residual, xu2019powerful} (refer to Appendix~\ref{subapp:timecomparison} for further results).
Since performance predictors aim to quickly identify effective neural architectures, this slower processing time poses a considerable practical bottleneck when using flow-based encoders.


Another line of research focuses on the effective pre-training of neural architecture encoders.
Neural architecture encoders require sufficient training data—architectures paired with their ground-truth performance on the target task and dataset—since deep learning models are often vulnerable to the overfitting issue.
However, obtaining enough neural architectures along with their performance entails heavy costs.
To address this, various \textit{generative pre-training methods} for neural architecture encoding have been explored~\citep{yan2020does, jing2022graph}. 
Note that in various domains, it is known that by using well-designed generative pretext tasks, deep learning models can capture data patterns~\citep{he2022masked, kenton2019bert, hu2019strategies, kim2024rethinking} that lead to performance improvements in label-scarce scenarios~\citep{zhang2023prompt, yuan2024spatio, kim2024hypeboy}.

\noindent\textbf{Challenge 2.}
However, existing generative pre-training methods,
which are often adapted from other domains~\citep{yan2020does, jing2022graph}, may be sub-optimal for neural architecture encoding.
Consider a generative pretext task that involves predicting masked parts of the input, used for both neural architectures~\citep{jing2022graph} and molecules~\citep{hu2019strategies}.
In Figure~\ref{fig:firstmotivation} (b), only a limited set of atoms can occupy the masked area due to chemical rules, which deep learning models learn through this task.
In contrast, such rules are largely absent in neural architectures; as shown in Figure~\ref{fig:firstmotivation} (b), most operations are possible options for the masked part. 
This characteristic makes it unclear what specific advantage the model gains from this generative pretext task in the context of neural architecture.

To address both key challenges, we propose a novel and specialized generative pre-training method for neural architecture encoding, \textbf{\method} (\textbf{\underline{F}}low-based \textbf{\underline{G}}enerative \textbf{\underline{P}}re-training). 
\method enables an encoder to capture the information flow within a neural architecture, even if the encoder structure is not specially designed to capture the information flow (e.g., the non-flow-based encoders~\citep{xu2019powerful, bresson2017residual}).
We train an encoder to
reconstruct \textit{flow surrogate}, a proposed representation of a neural architecture's information flow, allowing the encoder to capture the information flow of the architecture. 
\method addresses the two practical challenges outlined earlier: 
(1) it enables encoders to become flow-aware without requiring a specialized model structure that would increase computational time significantly, and 
(2) it provides clear training guidance to the encoder, allowing it to effectively learn the information flow within a neural architecture through our pretext task.

We demonstrate the effectiveness of our pre-training method compared to baseline pre-training methods across multiple downstream tasks, including performance prediction and neural architecture search.
Specifically, compared to the strongest baseline method~\citep{zhao2023dynamic}, \method shows up to 46.5\% gain in terms of Precision@1\% in performance prediction.
Our key contributions are as follows:

\begin{itemize}[leftmargin=*]
    \item We propose a novel generative pre-training method, \method, for neural architecture encoding. \method trains a neural architecture encoder to learn the information flow within a neural architecture even without a specialized encoder structure (Section~\ref{sec:method}).
    \item In the performance prediction experiment, \method outperforms the baseline pre-training methods in 23 out of 27 settings, demonstrating its efficacy (Section~\ref{subsec:performanceprediction}).
    \item In the neural architecture search (NAS) experiment—a key application of neural architecture encoders—\method outperforms the baseline pre-training methods (Section~\ref{subsec:nas}). 
\end{itemize}
\vspace{-1mm}
Our code and datasets are available at~\url{https://github.com/kswoo97/FGPAnom}.

\section{Related Work and Preliminaries}
\label{sec:relatedwork}
In this section, we first review related studies, followed by an outline of the preliminary concepts essential to our study.


\subsection{Related work}

\subsubsection{Neural architecture encoding.}
Neural architecture encoding aims to learn good representations of neural architectures~\citep{yi2023nar, hwang2024flowerformer, ning2022ta, ning2020generic, yi2023nar2}.
Its notable application is neural architecture performance prediction, which can reduce the cost of finding a proper neural architecture for the target task and dataset. 

Neural architectures, including Transformers~\citep{chitty2022neural} and convolutional neural networks~\citep{ying2019bench}, can be expressed as graphs where nodes represent operations (e.g., 3x3 conv and max pooling) and edges represent connections between them.
Therefore, graph neural networks (GNNs)~\citep{bresson2017residual, xu2019powerful, kipf2017semi} are widely used as neural architecture encoders~\citep{wei2022npenas, wen2020neural, yan2020does, chen2021contrastive, dudziak2020brp}.
GNNs encode a neural architecture graph by aggregating and transforming features from the neighborhood of each node. 

Recent works have increasingly focused on capturing the information flow within a given neural architecture by using specialized model structures~\cite{ning2020generic,ning2022ta, hwang2024flowerformer}.
A representative example is FlowerFormer~\citep{hwang2024flowerformer}, a state-of-the-art neural architecture encoder that employs sequential information processing to imitate the forward-pass and backpropagation in a neural architecture. 
Although more effective, these flow-based encoders require significantly more processing time, due to their sequential processing, compared to simpler GNN-based encoders (refer to Figure~\ref{fig:firstmotivation} (a)).



\subsubsection{Pre-training for neural architecture encoding.}
Due to the high cost of obtaining a neural architecture's ground-truth performance on the target task and dataset (i.e., the architecture's `label'), it is essential to train accurate performance predictors—typically using neural architecture encoders—in label-scarce scenarios, as discussed in Section~\ref{sec:introduction}.
To this end, generative pre-training methods~\citep{kipf2016variational, he2022masked}, known for enhancing the generalization capabilities of deep learning models~\citep{zhang2023prompt, yuan2024spatio, kim2024hypeboy}, have been adopted.
These methods leverage unlabeled neural architectures (i.e., those with unknown ground-truth performance) to pre-train a neural architecture encoder and fine-tune the encoder on the target downstream task, such as the performance prediction tasks~\citep{yan2020does, jing2022graph}.

Arch2vec~\cite{yan2020does} is based on a variational graph autoencoder~\citep{kipf2016variational}, which trains a neural architecture encoder to reconstruct edges in graphs representing neural architectures.
GMAE~\cite{jing2022graph} builds on masked autoencoder~\citep{he2022masked}, where certain operations within a neural architecture are masked, and the encoder is trained to predict these masked operations. 

In contrast, \citet{zhao2023dynamic} leveraged zero-cost proxies~\citep{abdelfattah2021zero, krishnakumar2022bench, huang2024up}, which are neural architecture's statistics correlated with its ground-truth performance.
They trained a neural architecture encoder to predict the zero-cost proxies of a neural architecture, enabling it to learn which types of architectures are more likely to yield high performance.

\subsection{Preliminary}
\label{subsec:prelim}

\begin{wrapfigure}[17]{r}{0.55\textwidth}  
  \centering
  \begin{minipage}[t]{0.2\textwidth}
    \includegraphics[width=\textwidth]{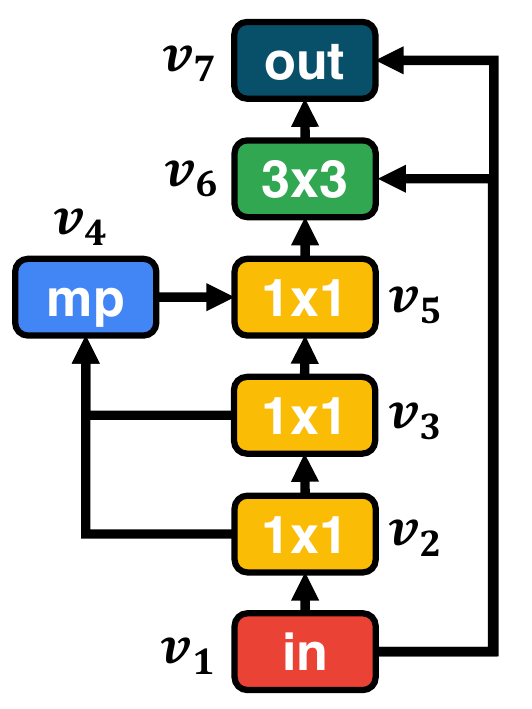}
    \small
    \textbf{(a)} An architecture from NAS-Bench-101.
    \label{fig:architecture}
  \end{minipage}
  \hspace{1mm}
  \begin{minipage}[t]{0.31\textwidth}
    \includegraphics[width=\textwidth]{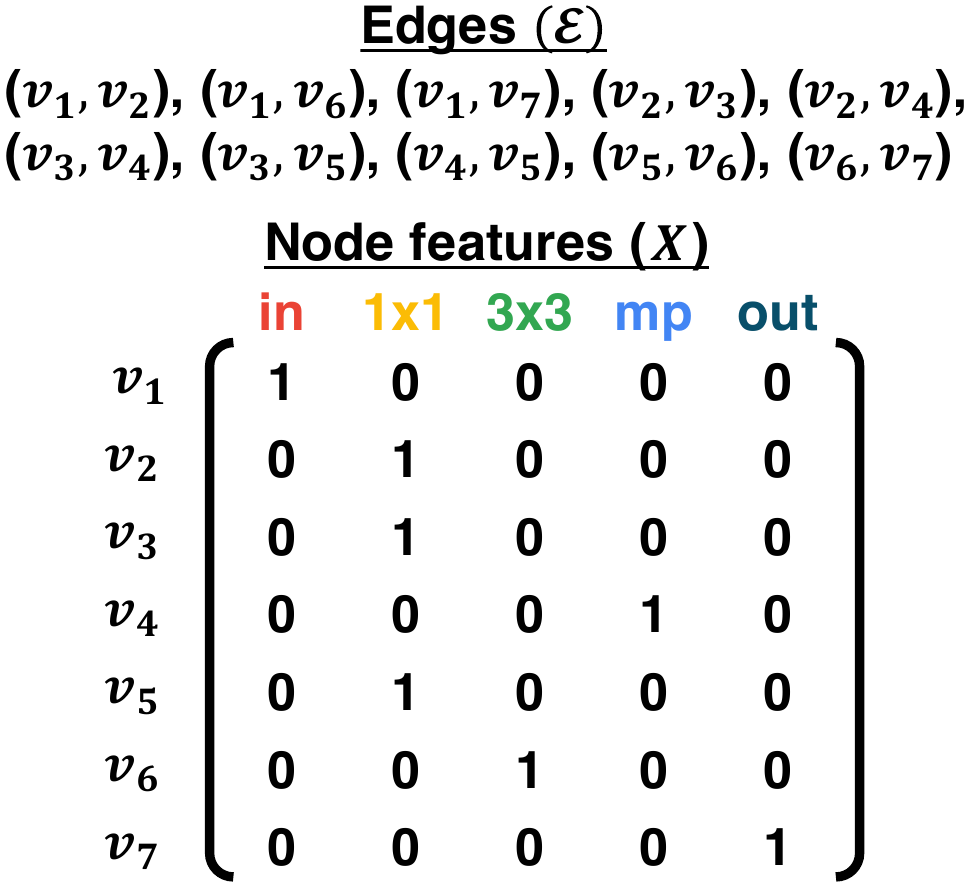}
    \small
    \textbf{(b)} Graph representation of the example neural architecture in \textbf{(a)}.
    \label{fig:graphrepresentation}
  \end{minipage}
  \caption{\textbf{Graph modeling of a neural architecture.} 
  An example of how a neural architecture is modeled as a directed acyclic graph.}
  \label{fig:representationexample}
\end{wrapfigure}

\textbf{Neural architecture graph.}
A neural architecture is often modeled as a directed acyclic graph $\mathcal{G} = (\mathcal{V}, \mathcal{E})$, defined by a set of nodes $\mathcal{V} = \{v_{1},\cdots,v_{\vert \mathcal{V}\vert}\}$ that represent operations, and a set of directed edges $\mathcal{E} \subseteq \{(v_{i},v_{j}) : v_{i},v_{j} \in \mathcal{V}\}$ that represent connections between consecutive operations (refer to Figure~\ref{fig:representationexample} for an example).
Specifically, when operation $v_{j}$ follows operation $v_{i}$, the two are connected by a directed edge $(v_{i},v_{j})$, with $v_{i}$ as the source and $v_{j}$ as the destination.
Each node is assigned a feature vector, typically a one-hot encoded vector indicating its operation type (refer to Figure~\ref{fig:representationexample}).
We denote the feature vector matrix as $\mathbf{X}\in \{0, 1\}^{\vert \mathcal{V}\vert \times \vert O\vert}$, where $\mathcal{O}$ represents the set of all possible operations.
The $i$-th row of $\mathbf{X}$, denoted by $\mathbf{X}_{i, :}$, corresponds to the feature vector of node $v_{i}$.
Therefore, a neural architecture can also be expressed as $\mathcal{G}= (\mathbf{X},\mathcal{E})$.

\noindent\textbf{Neural architecture encoders.}
Neural architecture encoders transform a neural architecture graph $\mathcal{G} = (\mathbf{X}, \mathcal{E})$ into its vector representation (i.e., embedding).
Formally, given $\mathcal{G}$ as an input, a neural architecture encoder $f_{\theta}$ produces a vector $\mathbf{z} \in \mathbb{R}^{d}$ (i.e., $f_{\theta}(\mathbf{X}, \mathcal{E}) = \mathbf{z}$).
For encoders that give embeddings for each operation~\citep{xu2019powerful, hwang2024flowerformer} (i.e., $f_{\theta}(\mathbf{X}, \mathcal{E}) = \mathbf{Z}' \in \mathbb{R}^{\vert \mathcal{V}\vert \times d}$), we apply mean pooling to obtain $\mathbf{z}$, as in~\cite{hwang2024flowerformer}.

\vspace{-2mm}
\section{Proposed method}
\label{sec:method}

In this section, we introduce our proposed generative pre-training method for neural architecture encoding, \textbf{\method} (\textbf{\underline{F}}low-based \textbf{\underline{G}}enerative \textbf{\underline{P}}re-training).
\method trains a neural architecture encoder to learn the information flow within a given neural architecture, even if the encoder lacks a specialized structure for capturing this flow. 

\subsection{Motivation, challenge, and overview}
\label{subsec:method_motivation}
\textbf{Motivation.} Information flow refers to how input data propagates through a neural network during the forward pass and how gradients are transmitted back during the backpropagation.
This concept plays a fundamental role in understanding how a neural network learns and makes predictions, making it a critical feature of neural architectures~\citep{ning2020generic,ning2022ta,hwang2024flowerformer}.
Although state-of-the-art neural architecture encoders successfully capture the information flow through specialized model structures~\citep{ning2022ta, hwang2024flowerformer}, complexity leads to significantly longer processing times compared to simpler, non-flow-based encoders (refer to Figure~\ref{fig:firstmotivation} (a)).
Therefore, we anticipate that training a simple neural architecture encoder—especially without a specialized structure for information flow—to learn this flow can address the computational efficiency issue while enhancing the encoder's overall effectiveness.
 

\noindent\textbf{Challenge.} 
The main challenge in learning the information flow is defining an appropriate \textit{training objective}.
To address this, we introduce \textit{flow surrogate}, a vector representing the information flow within a neural architecture.
This surrogate serves as a pre-training objective, guiding the encoder to understand the information flow in neural architectures.
We obtain the surrogate by passing random vectors through a directed acyclic graph that represents an architecture, which is a highly simplified simulation of a neural network's forward pass and backpropagation.\footnote{We analyze the alternatives and reliability of using random vectors in Appendix~\ref{subsec:usingrandom}.}
Notably, each architecture has its own flow surrogate, which can be obtained with a one-time computational effort without any learning procedure. 
This does not require prior knowledge of the architecture's ground-truth performance on the target task. 
We detail the process of obtaining the flow surrogate (i.e., pre-training objective) in Section~\ref{subsec:method_surrogate}. 

\noindent\textbf{Overview.} We now provide an overview of \method.
Given a set of unlabeled neural architectures (i.e., architectures with unknown ground-truth performance on the target task and dataset), we first express each architecture as a graph (i.e., neural architecture graph), as outlined in Section~\ref{subsec:prelim}.
Next, we obtain each architecture's flow surrogate (i.e., pre-training objective), as detailed in Section~\ref{subsec:method_surrogate}.
Lastly, we train a neural architecture encoder to learn the information flow by reconstructing the obtained flow surrogate, as detailed in Section~\ref{subsec:method_pipeline}.

\begin{figure*}[t]
  \centering
{\includegraphics[width=1.0\textwidth]{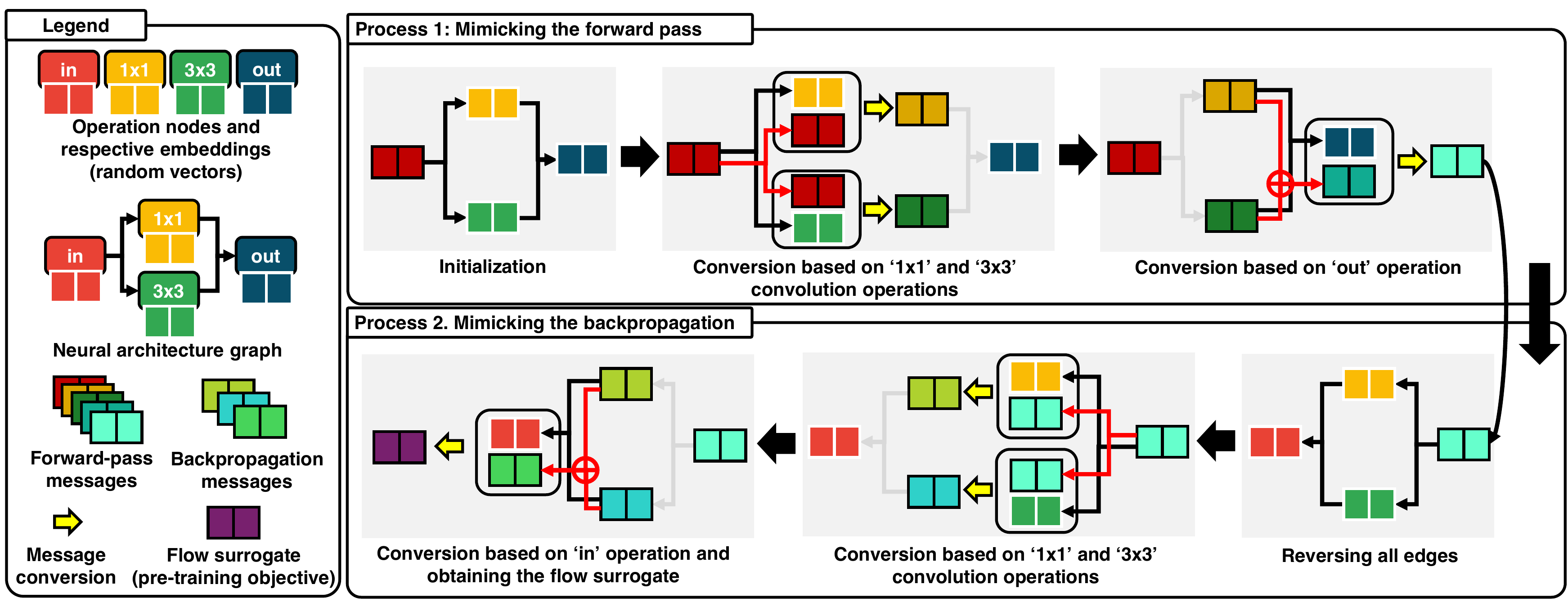}}
  \caption{\textbf{Generation of the proposed pre-training objective.} 
  Red edges mark active computation for the current step; gray edges indicate inactive computation.
  Starting from `in' (input), a message (spec., randomly-drawn embedding of `in') flows through the `1x1' and `3x3' operations, where it is converted depending on each operation type. The two converted messages are summed and flow to `out' (output) for further conversion.
 This process is then reversed, and the message flows back through the operations until it returns to `in'.
 The resulting message, called the architecture's flow surrogate, serves as the pre-training objective.}
 \vspace{3mm}
  \label{fig:surrogate}
\end{figure*}

\subsection{Proposed pre-training objective}
\label{subsec:method_surrogate}

We describe the process for obtaining the flow surrogate, our proposed representation of a neural architecture's information flow, which serves as our pre-training objective.
We sequentially propagate random vectors through a directed acyclic graph that represents a neural architecture. 
Specifically, the random vectors, called messages, are passed between nodes (operations) along edges (connections between operations).
When a node receives messages from other nodes, its message is converted based on its specific operation type, capturing unique operational characteristics.
Notably, to simulate both the forward pass and backpropagation in a highly simplified manner, we: (1) propagate messages in the forward direction along the edges to model the forward pass, and (2) repeat the process in the reverse direction to model backpropagation.
An overview is in Figure~\ref{fig:surrogate}.
After completing both propagation steps, the final converted message(s) serve as the flow surrogate for the corresponding neural architecture.
We detail this process in three steps: (1) assigning a topological order, (2) mimicking the forward pass, and (3) mimicking the backpropagation.
Note that this process resembles how flow-based encoders generate outputs. 
However, in our approach, the process is used only \textbf{once} per architecture (with random vectors) to obtain pre-training objectives and not repeated for training.

\begin{wrapfigure}[11]{r}{0.4\textwidth}  
  \vspace{-5mm}  
  \centering
  \includegraphics[width=\linewidth]{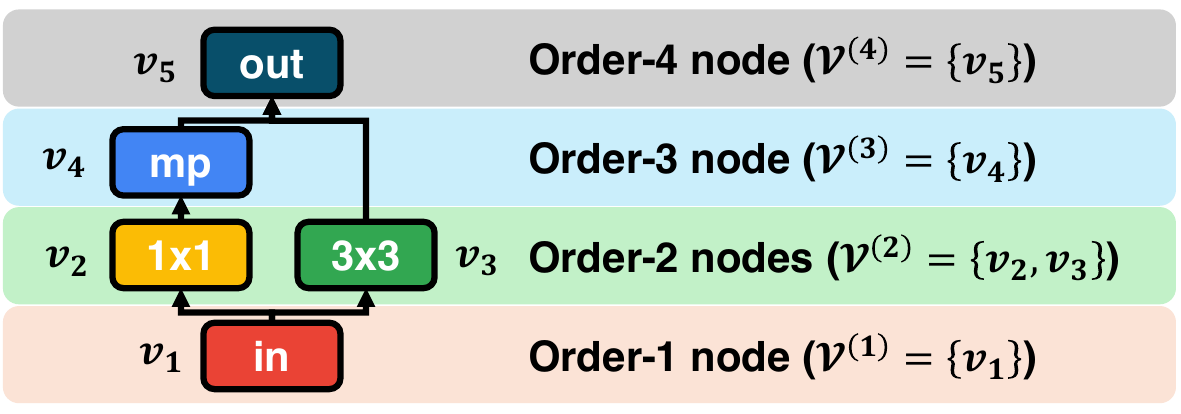}
  \caption{\textbf{Example of topological order assignment in a neural architecture graph.}
  Assignment results are ($\mathcal{V}^{(1)} = \{v_{1}\}$, $\mathcal{V}^{(2)} = \{v_{2}, v_{3}\}$, 
  $\mathcal{V}^{(3)} = \{v_{4}\}$, and $\mathcal{V}^{(4)} = \{v_{5}\}$), and $T = 4$.}
  \label{fig:topologicalorder}
  \vspace{-5mm}  
\end{wrapfigure}

\subsubsection{Assigning topological order}
\label{subsec:orderassignment}
For a given neural architecture, we first obtain its graph expression $\mathcal{G} = (\mathbf{X}, \mathcal{E})$, as outlined in Section~\ref{subsec:prelim}.
Recall that this graph is a directed acyclic graph. 
In this graph, we assign a topological order to each node, indicating the sequence in which each node receives the information flow.
An example of topological order assignment is provided in Figure~\ref{fig:topologicalorder}.
{Order-1} nodes are defined as nodes without any incoming edges. 
After finding order-1 nodes, we remove them and the edges where they serve as source nodes from the graph.
Next, {order-2} nodes are those that, in the modified graph, have no incoming edges.
We repeat this process until all nodes have been assigned an order.
We denote a set of order-$t$ nodes as $\mathcal{V}^{(t)}$, and denote the last order as $T$. 
Thus, the node set $\mathcal{V}$ is split into disjoint subsets $\mathcal{V}^{(1)}, \dots , \mathcal{V}^{(T)}$.

\subsubsection{Mimicking the forward pass}
\label{sec:forwardpass_mimicking}

Once the topological order assignment (Section~\ref{subsec:orderassignment}) is complete, we model the forward pass of a neural architecture.
To this end, we propagate forward-pass messages from node to node along the edges of the neural architecture graph.
Here, the forward-pass message of a node represents the hidden state at the corresponding node (i.e., operation) within the neural architecture, and each node has a distinct forward-pass message.
We refer to these forward-pass messages as \textit{fp-messages}.
When a node $v_{i}$ receives fp-messages from other nodes along edges, the fp-message of $v_{i}$ is converted according to the incoming fp-messages and the operation type of $v_{i}$.
The process of propagation and conversion proceeds in the order of $\mathcal{V}^{(1)}, \mathcal{V}^{(2)}, \dots, \mathcal{V}^{(T)}$, until the fp-messages of order-$T$ nodes (i.e., $v_{j} \in \mathcal{V}^{(T)}$) are converted.
An example is provided in the `Process 1' box of Figure~\ref{fig:surrogate}.

We denote a fp-message of a node $v_{i}$ as $\mathbf{f}_{i} \in \mathbb{R}^{k}$.
Also, each node $v_{i} \in \mathcal{V}$ has an embedding, denoted by $\mathbf{h}_{i} \in \mathbb{R}^{k}$, based on its operation type.
Specifically, by using a matrix $\mathbf{P} \in \mathbb{R}^{\vert \mathcal{O}\vert \times k}$ randomly sampled from $\mathcal{N}(0,\sigma^{2})$, we compute the node embedding matrix $\mathbf{H}=\mathbf{PX}$, where its $i-$th row corresponds to $v_{i}$'s embedding (i.e., $\mathbf{H}_{i,:} = \mathbf{h}_{i}$).
Note that nodes performing the same operation share the same embedding across all architectures.
Additionally, we set the fp-messages of the order-1 nodes, which do not have any incoming edges, with a \textit{randomly initialized vector} $\mathbf{r} \in \mathbb{R}^{k}$ (i.e., $\mathbf{f}_{j} = \mathbf{r}, \forall v_{j} \in \mathcal{V}^{(1)}$).
Note that fp-messages of all order-1 nodes are initialized as $\mathbf{r}$ in every neural architecture.

We formalize the propagation and conversion process at node $v_{i} \in \mathcal{V}^{(t)}$.
When multiple fp-messages arrive at $v_{i}$ (e.g., `out' operation node in Figure~\ref{fig:surrogate}), we first aggregate these arriving fp-messages using sum pooling as
$\mathbf{m}_{i} = \sum_{v_{j} \in \mathcal{N}^{(i)}} \mathbf{f}_{j}$, where $  \mathcal{N}^{(i)} = \{v_{j} : (v_{j},v_{i}) \in \mathcal{E}\}.$
In this process, using mean or max (rather than sum) pooling for neighbor aggregation yields lower performance than our approach ({refer to} Appendix~\ref{subapp:pooling}).

Next, we convert the pooled fp-messages $\mathbf{m}_{i} \in \mathbb{R}^{k}$ according to the operation of $v_{i}$, obtaining the converted fp-message of $v_{i}$, denoted by $\mathbf{f}_{i} \in \mathbb{R}^{k}$, as
$\mathbf{f}_{i} = \alpha \mathbf{m}_{i} + (1 - \alpha)\texttt{ReLU}\left([\mathbf{h}_{i} \Vert \mathbf{m}_{i}]\mathbf{W}\right)$, 
where $[\mathbf{a} \Vert \mathbf{b}]$ is a concatenation of vectors $\mathbf{a}$ and $\mathbf{b}$, $\mathbf{W} \in \mathbb{R}^{2k \times k}$ is a fixed projection matrix, and $\alpha \in [0, 1]$ is a weighing hyperparameter.
Here, $\mathbf{W}$ and $\alpha$ are shared across all architectures and nodes.
When the fp-messages of order-$T$ nodes (i.e., $v_{j} \in \mathcal{V}^{(T)}$) are converted, the process of mimicking the forward pass is complete.

\subsubsection{Mimicking the backpropagation}
\label{subsec:backpropagation_mimicking}

Once the process of mimicking the forward pass (Section~\ref{sec:forwardpass_mimicking}) is done, we do a reverse propagation process. 
This process aims to mimic the backpropagation step in a neural network.
To this end, we propagate backpropagation messages, which we call \textit{bp-messages}. 
The bp-message of a node represents the gradient at the corresponding node (i.e., operation) within the neural architecture, and each node has a distinct bp-message.
The overall process (i.e., propagation and conversion) is similar to that for mimicking the forward pass (Section~\ref{sec:forwardpass_mimicking}), but this process proceeds in the order of $\mathcal{V}^{(T)}, \mathcal{V}^{(T-1)}, \dots, \mathcal{V}^{(1)}$, until the bp-messages of order-$1$ nodes (i.e., $v_{j} \in \mathcal{V}^{(1)}$) are converted.
An example is in the `Process 2' box of Figure~\ref{fig:surrogate}.
We denote the bp-message of node $v_{i}$ as $\mathbf{b}_{i} \in \mathbb{R}^{k}$.
Furthermore, each order-$T$ node's bp-message is initialized to the corresponding node's fp-message (i.e., $\mathbf{b}_{j}=\mathbf{f}_{j}, \forall v_{j} \in \mathcal{V}^{(T)}$), to mimic the dependence of backpropagation on the forward pass.

We formalize the propagation and conversion process at node $v_{i} \in \mathcal{V}^{(t)}$.
Here, bp-messages arrive from the nodes at the endpoints of $v_{i}$'s outgoing edges, and when multiple bp-messages arrive at $v_{i}$ (e.g., `in' operation node in Figure~\ref{fig:surrogate}), we aggregate these incoming bp-messages using sum pooling as follows:
$\mathbf{m}'_{i} = \sum_{v_{j} \in \mathcal{K}^{(i)}} \mathbf{b}_{j},$ where $  \mathcal{K}^{(i)} = \{v_{j} : (v_{i},v_{j}) \in \mathcal{E}\}.$

We then convert the pooled bp-messages $\mathbf{m}'_{i} \in \mathbb{R}^{k}$ according to the operation of $v_{i}$, obtaining the converted bp-message of $v_{i}$, denoted by $\mathbf{b}_{i} \in \mathbb{R}^{k}$, as $\mathbf{b}_{i} = \alpha \mathbf{m}'_{i} + (1 - \alpha)\texttt{ReLU}\left([\mathbf{h}_{i} \Vert \mathbf{m}'_{i}]\mathbf{W}\right)$.
When the bp-messages of order-$1$ nodes (i.e., $v_{j} \in \mathcal{V}^{(1)}$) are converted, the process of mimicking the backpropagation is complete.

After mimicking the forward-pass and backpropagation, we sum the bp-messages of the order-1 nodes to obtain the flow surrogate $\mathbf{s} \in \mathbb{R}^{k}$ (i.e., $\mathbf{s} = \sum_{v_{i} \in \mathcal{V}^{(1)}} \mathbf{b}_{i}$).
The resulting vector $\mathbf{s}$ serves as the pre-training objective for the corresponding architecture, the flow surrogate representing the information flow within $\mathcal{G}$.
Although this process can be repeated over multiple rounds, such methods underperform compared to our approach, as detailed in Appendix~\ref{subapp:newdesign}.
Further discussion of the surrogate’s expressiveness and theoretical properties is in Appendices~\ref{subapp:expressiveness} and~\ref{subapp:theory}, respectively.

\subsection{Proposed flow generative pre-training}
\label{subsec:method_pipeline}
We now present the details of \method, our flow generative pre-training method for neural architecture encoding. 
As outlined in Section~\ref{subsec:method_motivation}, \method trains a neural architecture encoder to reconstruct the flow surrogate (i.e., pre-training objective) $\mathbf{s} \in \mathbb{R}^{k}$ (Section~\ref{subsec:method_surrogate}) of a given neural architecture (Section~\ref{subsec:method_surrogate}).
Then, \textit{any} encoder, without requiring a specialized model structure, is trained to
capture the information flow 
through three steps: (1) encoding, (2) decoding, and (3) computing the pre-training loss.

\noindent\textbf{Encoding and decoding.}
For a given neural architecture, we start by deriving its graph expression $\mathcal{G} = (\mathbf{X}, \mathcal{E})$, as described in Section~\ref{subsec:prelim}. 
We then encode this graph into an embedding $\mathbf{z} \in \mathbb{R}^{d}$ using a neural architecture encoder $f_{\theta}$ (i.e., $f_{\theta}(\mathbf{X}, \mathcal{E}) = \mathbf{z}$), which we aim to train.
After, we decode the embedding by using an MLP decoder $g_{\phi}$ to obtain the reconstructed surrogate $\hat{\mathbf{s}} \in \mathbb{R}^{k}$ (i.e., $g_{\phi}(\mathbf{z}) = \hat{\mathbf{s}}$), ensuring the encoder focuses on learning generalizable representations while the decoder handles task-specific transformations.

\noindent\textbf{Pre-training loss.}
Then, we compute the reconstruction loss $\mathcal{L}_{rec}$ by measuring the squared $\ell_{2}$-distance between the original surrogate $\mathbf{s}$ (Section~\ref{subsec:method_surrogate}) and the reconstructed surrogate $\hat{\mathbf{s}}$ (i.e., $\mathcal{L}_{rec} = \lVert \mathbf{s} - \hat{\mathbf{s}}\rVert^{2}_{2}$).
Notably, \method can be combined with auxiliary learning objectives, such as predicting zero-cost proxies of a neural architecture~\citep{abdelfattah2021zero, lee2024az}.
The overall training loss is $\mathcal{L} = \lambda_{1}\mathcal{L}_{rec} + \lambda_{2}\mathcal{L}_{aux}$, where $\lambda_{1}$ and $\lambda_{2}$ are loss-weighing hyperparameters.
Details on our usage of auxiliary learning objectives are provided in Appendix~\ref{sec:zcusage}.
The learnable parameters $\theta$ and $\phi$ are optimized to minimize $\mathcal{L}$ via gradient descent.
After pre-training, the trained encoder $f_{\theta}$ can be fine-tuned to perform certain downstream tasks, such as neural architecture's performance prediction.

\vspace{-2mm}
\section{Experiment}
\vspace{-2mm}
\label{sec:experiment}

In this section, we demonstrate the effectiveness of \method in several applications, including performance prediction and neural architecture search. 
Specifically, we answer the following questions: 
\begin{itemize}[leftmargin=*]
    \item RQ1. How effective is \method in predicting the performance of neural architectures?
    \item RQ2. How effective is \method in neural architecture search?
    \item RQ3. Can the proposed flow surrogate well represent the ground-truth performance of architectures?
    \item RQ4. Are all the key components of \method essential? 
    \item RQ5. How long does it take to train an encoder by \method?
\end{itemize}



\subsection{Experimental setup}


{\textbf{Datasets and splits.}}
We leverage three computer vision neural architecture datasets, which are NAS-Bench-101 (NB-101)~\citep{ying2019bench}, NAS-Bench-201 (NB-201)~\citep{dong2020bench}, and NAS-Bench-301 (NB-301)~\citep{siems2020bench} datasets.
In addition, we provide results on other domain datasets (spec., natural language processing and graph representation learning) in Appendix~\ref{subapp:newdomain}. 
For NB-101 and NB-201 datasets, we follow the training and test splits provided in~\citep{ning2022ta,hwang2024flowerformer}.
For the NB-301 dataset, since the baseline method ZC-Proxy~\citep{zhao2023dynamic} requires certain numerical properties of architectures, we use a subset of the original NB-301 dataset where these properties are available.
We sample 40 architectures from the test set to create a validation set, following the approach in~\citep{hwang2024flowerformer}.
Further details are provided in Appendix~\ref{subapp:dataset}.

\textbf{Backbone encoders.}
We use 3 backbone graph-based neural architecture encoders, ResGatedGCN~\citep{bresson2017residual}, GIN~\citep{xu2019powerful}, and FlowerFormer~\citep{hwang2024flowerformer}.
Specifically, ResGatedGCN and GIN are GNN-based encoders, which do not have a specialized structure for capturing the information flow (i.e., non-flow-based encoders).
In contrast, FlowerFormer is the SOTA flow-based encoder with a specialized design to capture the flow.
Moreover, we present analysis for non-graph-based models in Appendix~\ref{subapp:backbone}.

\textbf{Baseline methods.}
We use five baselines: an encoder trained solely on supervised learning (N/A) and four pre-training methods for neural architecture encoding, which are the (1) graph-contrastive-learning method (GraphCL~\citep{you2020graph}) and (2) generative methods based on connections (Arch2vec~\citep{yan2020does}), operations (GMAE~\citep{jing2022graph}), and zero-cost proxies (ZC-Proxy~\citep{zhao2023dynamic}).

\textbf{Pre-training, fine-tuning, and evaluation protocol.}
For each dataset and neural architecture encoder, we first pre-train the encoder using a specific pre-training method across the whole dataset (i.e., both training and test sets).
Note that ground-truth performances of any architectures are \textbf{not used} during this pre-training stage, and despite this fact, it is also possible to use only the training set even for pre-training (see Appendix~\ref{appendix:strict}).
Following pre-training, we fine-tune the encoder using only the training set to optimize it for the performance prediction task, using the ground-truth performance of the training set in this phase.
After fine-tuning, we evaluate the encoder on the test set using three evaluation metrics: Kendall's tau~\citep{sen1968estimates}, Precision-@1\%, and Precision-@5\%, all widely adopted for this evaluation process~\citep{ning2022ta, hwang2024flowerformer}.
We employ three distinct dataset splits and three model initializations, resulting in a total of nine experimental configurations. 
Further details are in Appendix~\ref{app:expdetail}.

\begin{table*}[t]

\caption{\textbf{Performance prediction results.}
Mean and standard deviation on each dataset and metric.
The best performance is highlighted in \colorbox{sunwoogreen}{blue}.
N/A denotes an encoder solely trained with supervised learning, without any pre-training method.
Gain from N/A denotes the performance improvement of \method compared to N/A.
{All values are multiplied by 100 {to save space in the table}.}
\method outperforms the baseline pre-training methods in 23 out of 27 settings.}
\setlength{\tabcolsep}{5.0pt}
\small
\centering
\scalebox{0.78}{
\renewcommand{\arraystretch}{1.0}
\begin{tabular}{c | c | c c c | c c c | c c c}
    \toprule
    \multirow{2}{*}{\makecell{Encoder}} & \multirow{2}{*}{\makecell{Pre-training \\ method}} & \multicolumn{3}{c|}{Kendall's Tau} 
    & \multicolumn{3}{c|}{Precision@1\%} & \multicolumn{3}{c}{Precision@5\%} \\
    
    
         &  & NB-101 & NB-201 & NB-301
            & NB-101 & NB-201 & NB-301
            & NB-101 & NB-201 & NB-301 
            \\
    \midrule 
    \midrule 
    
    \multirow{7}{*}{ResGatedGCN}
     
     & N/A & 65.0 {\std (7.8)} & 73.4 {\std (1.5)} & 54.4 {\std (4.1)}
     & 18.2 {\std (7.9)} & 29.7 {\std (11.8)} & 18.4 {\std (2.6)}
     & 46.2 {\std (12.2)} & 51.7 {\std (4.2)} & 40.6 {\std (1.7)}
     \\
     
     & GraphCL
     & 66.9 {\std (5.0)} & 73.7 {\std (1.9)} & 54.6 {\std (3.5)} 
     & 21.3 {\std (8.8)} & 31.6 {\std (12.2)} & 20.4 {\std (1.7)} 
     & 46.5 {\std (8.3)} & 52.3 {\std (5.6)} & 39.7 {\std (2.3)} 
     \\
     
     & Arch2vec
     & 65.8 {\std (5.9)} & 74.1 {\std (1.2)} & 57.7 {\std (3.0)} 
     & 21.8 {\std (13.4)} & 28.3 {\std (11.3)} & \best 25.5 {\std (3.0)} 
     & 44.9 {\std (9.8)} & 52.5 {\std (5.1)} & 42.9 {\std (3.0)}
     \\

     & GMAE
     & 68.1 {\std (4.7)} & 74.8 {\std (1.2)} & 57.0 {\std (2.7)} 
     & 21.8 {\std (11.7)} & 35.1 {\std (11.2)} & 22.2 {\std (5.1)}
     & 49.9 {\std (6.3)} & 57.8 {\std (5.4)} & 41.8 {\std (3.5)} 
     \\

     & ZC-Proxy
     &  68.3 {\std (6.7)} &  79.9 {\std (0.8)} &   57.9 {\std (2.5)}
     &  26.2 {\std (6.2)} & 44.3 {\std (11.0)} & 23.0 {\std (3.1)}
     &  52.1 {\std (6.3)} & 61.5 {\std (3.0)} & 42.6 {\std (3.2)}
     \\

     \cmidrule{2-11}

     & \method 
     & \best 74.8 {\std (4.8)} & \best 82.2 {\std (0.7)} & \best 58.4 {\std (2.4)}  
     & \best 37.5 {\std (13.0)} & \best 48.9 {\std (8.1)} &  23.2 {\std (4.6)} 
     & \best 61.7 {\std (3.6)} & \best 62.3 {\std (2.5)} & \best 43.2 {\std (2.3)} 
     \\

     & Gain from N/A & +15.1\% & +12.0\% & +7.4\%
     & +106.0\% & +64.6\% & +26.1\% 
     & +33.6\% & +20.5\% & +6.4\% \\
     
     \midrule 
     \midrule 

     \multirow{7}{*}{GIN}
     
     & N/A 
     & 62.8 {\std (5.9)} & 65.7 {\std (1.4)} & 52.3 {\std (2.7)} 
     & 26.9 {\std (14.9)} & 25.0 {\std (11.8)} & 19.2 {\std (3.5)} 
     & 48.1 {\std (12.0)} & 47.6 {\std (5.5)} & 38.9 {\std (1.3)}
     \\
     
     & GraphCL
     & 64.9 {\std (3.8)} & 66.9 {\std (0.9)} & 52.4 {\std (2.8)}
     & 32.3 {\std (18.9)} & 18.9 {\std (6.2)} & 20.2 {\std (3.4)}
     & 50.2 {\std (6.3)} & 45.0 {\std (3.0)} & 39.6 {\std (1.4)}
     \\
     
     & Arch2vec
     & 63.7 {\std (2.5)} & 68.0 {\std (0.7)} & 55.0 {\std (1.7)}
     & 30.0 {\std (19.0)} & 18.6 {\std (7.4)} &  21.2 {\std (3.1)}
     & 49.4 {\std (6.4)} & 45.6 {\std (2.8)} & 40.7 {\std (1.7)}
     \\

     & GMAE
     & 66.4 {\std (4.2)} & 70.6 {\std (2.2)} &  \best 55.3 {\std (1.6)}
     & 31.0 {\std (13.4)} & 27.8 {\std (14.4)} & 20.8 {\std (3.1)}
     & 52.9 {\std (3.0)} & 49.7 {\std (5.0)} &  41.5 {\std (1.9)}
     \\

     & ZC-Proxy
     & 65.2 {\std (8.3)} &  75.3 {\std (2.3)} & 53.0 {\std (3.0)}
     & 22.7 {\std (12.4)} & 29.0 {\std (14.3)} & 18.2 {\std (4.4)}
     & 49.9 {\std (13.0)} & 50.4 {\std (7.2)} & 38.2 {\std (2.6)}
     \\

     \cmidrule{2-11}

     & \method 
     & \best 67.8 {\std (3.7)} & \best 79.2 {\std (1.7)} & 55.2 {\std (2.2)}
     & \best 33.2 {\std (10.0)} & \best 35.6 {\std (12.7)} & \best 23.5 {\std (2.2)}
     & \best 55.8 {\std (3.8)} & \best 54.1 {\std (3.6)} & \best 42.0 {\std (2.2)}
    \\

    & Gain from N/A & +8.0\% & +20.6\% & +5.5\%
     & +23.4\% & +42.4\% & +22.4\% 
     & +16.0\% & +13.7\% & +8.0\% 
     \\
    
    \midrule 
     \midrule 

     \multirow{7}{*}{FlowerFormer}
     
     & N/A 
     & 74.0 {\std (3.6)} & 77.3 {\std (1.5)} & 55.6 {\std (4.7)}
     & 35.3 {\std (13.9)} & 35.6 {\std (14.1)} & 19.4 {\std (4.5)}
     & 56.4 {\std (5.2)} & 56.2 {\std (4.3)} & 40.7 {\std (4.2)}
     \\
     
     & GraphCL
     & 69.4 {\std (4.7)} & 77.0 {\std (3.2)} & 55.7 {\std (3.9)}
     & 35.3 {\std (7.7)} & 34.1 {\std (14.9)} & 21.2 {\std (2.6)}
     & 57.5 {\std (2.5)} & 55.7 {\std (6.3)} & 41.9 {\std (1.9)}
     \\
     
     & Arch2vec
     &  76.0 {\std (2.8)} & 77.8 {\std (1.8)} &  59.2 {\std (2.9)}
     & 38.4 {\std (9.1)} & 33.3 {\std (15.0)} & 24.2 {\std (3.0)}
     & \best 59.8 {\std (3.9)} & 57.5 {\std (6.0)} & 44.3 {\std (2.6)}
     \\

     & GMAE
     & 74.3 {\std (3.2)} & 78.7 {\std (1.1)} & 58.9 {\std (3.2)}
     & 33.6 {\std (9.6)} & 36.4 {\std (16.2)} & \best 26.3 {\std (5.2)}
     & 56.2 {\std (5.0)} & 58.7 {\std (5.9)} & 44.5 {\std (1.8)}
     \\

     & ZC-Proxy
     & 74.6 {\std (3.9)} &  82.3 {\std (1.2)} & 58.5 {\std (2.2)}
     &  37.8 {\std (5.7)} &  45.0 {\std (6.6)} & 25.3 {\std (2.9)}
     & 58.1 {\std (3.3)} & 60.8 {\std (3.8)} & 41.9 {\std (1.8)}
     \\

     \cmidrule{2-11}

     & \method 
     & \best 76.3 {\std (3.6)} & \best 83.6 {\std (1.7)} & \best 60.1 {\std (2.9)}
     & \best 40.6 {\std (13.1)} & \best 48.3 {\std (3.3)} & 24.2 {\std (5.4)}
     & 58.9 {\std (4.0)} & \best 65.0 {\std (4.5)} & \best 45.0 {\std (1.4)}
     \\

     & Gain from N/A & +3.1\% & +8.2\% & +8.1\%
     & +15.0\% & +35.7\% & +24.7\% 
     & +4.4\% & +15.7\% & +10.6\% \\
     
     \bottomrule
\end{tabular}
\label{tab:mainexp}
}
\vspace{3mm}
\end{table*}

\subsection{RQ1: Performance prediction experiments}
\label{subsec:performanceprediction}

We assess \method in the performance prediction task compared to the baseline pre-training methods.

\textbf{Setup.}
In practice, only a small subset of the full neural architecture search space has known ground-truth performance values due to the high cost of training and evaluating architectures. 
To simulate this circumstance, we use 1\% of the training set for fine-tuning neural architecture encoders—a common setting in neural architecture encoding research~\citep{ning2022ta, hwang2024flowerformer}.
In addition, analyses under varying (1) pre-training dataset sizes and (2) training dataset sizes are in Appendix~\ref{subapp:differentpretrainingdatasize} and \ref{subapp:additionalratio}, respectively.

\textbf{Results.}
As shown in Table~\ref{tab:mainexp}, \method outperforms all baselines in 23 out of 27 settings, demonstrating the effectiveness of learning information flow through pre-training in performance prediction.
Note that the performance improvement achieved by \method in the performance prediction task is not restricted to a specific dataset or encoder.
While the improvement is greater in the non-flow-based encoders (i.e., ResGatedGCN and GIN), it can also improve the performance of the flow-based encoder (i.e., FlowerFormer); we further explore the potential reasons behind the improvement of FlowerFormer in Appendix~\ref{subapp:flowerformergain}.

\subsection{RQ2: Neural architecture search experiments}
\label{subsec:nas}

Performance prediction is often integrated into neural architecture search (NAS) to automatically identify neural architectures suitable to the target task and dataset.
Accordingly, we evaluate the efficacy of \method in NAS, assessing its performance against those of existing pre-training methods.

\textbf{Setup.} 
As in \cite{hwang2024flowerformer}, we use NPENAS~\citep{wei2022npenas} as our backbone NAS algorithm, which uses a performance predictor to evaluate neural architectures in the search process.
Note that an accurate performance predictor enhances search efficacy, leading to the discovery of higher-performing architectures. 
\begin{wrapfigure}[15]{r}{0.45\textwidth}  
  \centering
  \includegraphics[width=\linewidth]{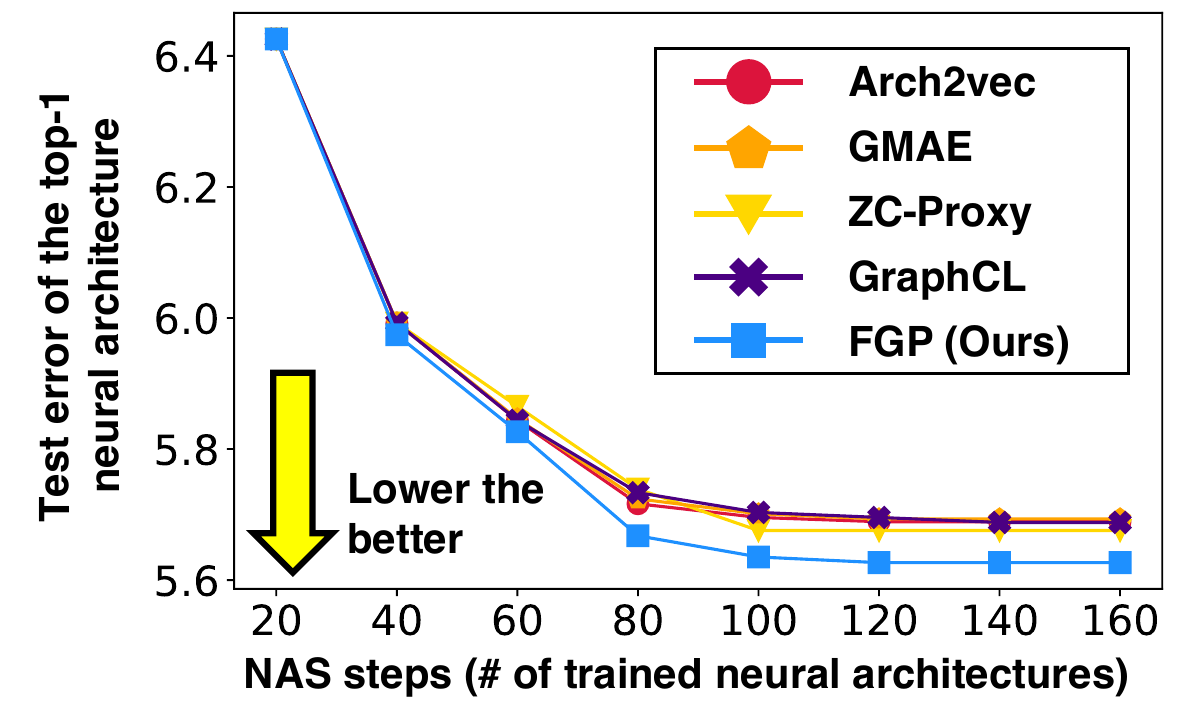}
  \caption{\textbf{NAS results.} The mean test error of the top-1 architecture across 10 trials is reported.
  The backbone NAS algorithm NPENAS~\citep{wei2022npenas} adopts pre-trained performance predictors, with each line representing a distinct pre-training approach.}
  \label{fig:nas} %
\end{wrapfigure}
To assess each pre-training method, we (1) pre-train a performance predictor on neural architectures within the designated search space and (2) initialize NPENAS’s predictor with this pre-trained model at the beginning of each search.
We run 10 trials with different initializations.
NAS-Bench-201 and ResGatedGCN serve as the search space and performance predictor, respectively.
Further details are in Appendix~\ref{subapp:nasdetail}.

\textbf{Results.}
As shown in Figure~\ref{fig:nas}, the NAS method, equipped with a performance predictor pre-trained using \method, identifies the more effective neural architecture than all competitors, each employing a distinct pre-training method.
Notably, \method consistently performs best at every NAS step. 

\begin{wrapfigure}[12]{r}{0.55\textwidth}  
  \small
  \vspace{-8mm}  
  \centering
  \includegraphics[width=\linewidth]{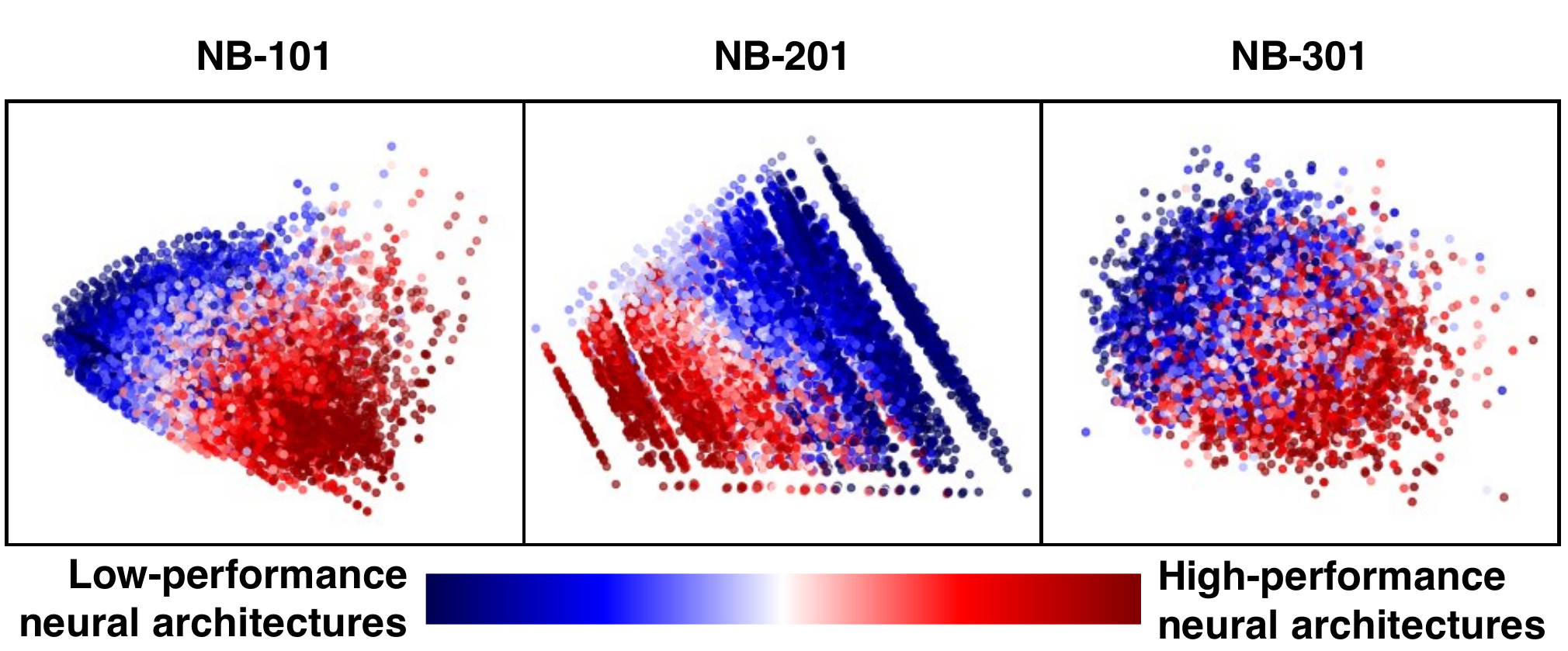}
  \caption{\textbf{Pre-training objective visualization.} PCA visualization of flow surrogates (i.e., pre-training objective), where colors represent the performance of the corresponding architecture.}
  \label{fig:pca}
  \vspace{3mm}  
\end{wrapfigure}

\subsection{RQ3: Flow surrogate analysis}
We provide a qualitative analysis of the relation between the flow surrogate of neural architecture (Section~\ref{subsec:method_surrogate}) and the ground-truth performance of the corresponding architecture.
Specifically, we aim to verify whether our flow surrogate well represents the architecture's ground-truth performance, separating architectures with differing performance..

\textbf{Setup.}
We visualize the distribution of the flow surrogate by using PCA.
We color each data point according to the corresponding neural architecture's ground-truth performance. 

\textbf{Results.}
As shown in Figure~\ref{fig:pca}, the proposed flow surrogate effectively represents the performance of neural architectures, distinguishing high-performing architectures from low-performing ones in separate regions. 
This visualization indicates that our pre-training objective, flow surrogate, can well guide a neural architecture encoder in identifying architectures likely to achieve high performance.

\vspace{-2mm}

\subsection{RQ4: Ablation study}

We demonstrate the necessity of the key components of \method in achieving high performance.


     
    



    
     

\begin{wrapfigure}[7]{r}{0.55\textwidth}  
\vspace{-5mm}
\centering
\caption{\textbf{Ablation study.}
Performance of the variants of \method.
All values are multiplied by 100 {to save space}.
The best performance is highlighted in \colorbox{sunwoogreen}{blue}.
}
\begin{varwidth}{\linewidth}
\small
\setlength{\tabcolsep}{2.0pt}
\scalebox{0.77}{
\renewcommand{\arraystretch}{1.1}
\begin{tabular}{l | c c c | c c c}
    \toprule
    \multirow{2}{*}{\makecell{Variants \\ of \method}} & 
    \multicolumn{3}{c|}{Kendall's Tau} 
    & \multicolumn{3}{c}{Precision@1\%} \\
    & NB-101 & NB-201 & NB-301 & NB-101 & NB-201 & NB-301\\
    \midrule \midrule 
     w/o $\mathcal{L}_{gen}$ 
     & 68.3 {\std (6.7)} & 79.9 {\std (0.8)} & 57.9 {\std (2.5)}
     & 26.2 {\std (6.2)} & 44.3 {\std (11.0)} & 23.0 {\std (3.1)} \\
     w/o $\mathcal{L}_{aux}$ 
     & 71.5 {\std (5.6)} & 74.5 {\std (0.7)} & 58.0 {\std (2.7)}
     & 35.6 {\std (8.7)} & 36.5 {\std (2.2)} & 23.6 {\std (3.2)} \\
     w/o Forward 
     & 73.5 {\std (4.3)} &  81.4 {\std (0.7)} & 57.1 {\std (3.6)}
     & 36.3 {\std (9.2)} &  44.6 {\std (9.0)} & 23.4 {\std (4.1)} \\
     w/o Backward 
     & 72.4 {\std (4.8)} &  81.4 {\std (0.6)} & 57.5 {\std (2.5)}
     & 31.0 {\std (9.8)} &  43.3 {\std (9.1)} & \best 24.2 {\std (2.7)} \\
    \midrule 
     \method 
     &  \best 74.8 {\std (4.8)} & \best 82.2 {\std (0.7)} & \best 58.4 {\std (2.4)}
     & \best 37.5 {\std (13.0)} & \best 48.9 {\std (8.1)} & 23.2 {\std (4.6)} \\
    \bottomrule
\end{tabular}
}
\end{varwidth}
\label{tab:ablexp}
\end{wrapfigure}

\textbf{Setup.}
We use four variants of \method: 
\begin{itemize}[leftmargin=*]
    \item \textbf{V1.} This does not perform the flow surrogate reconstruction (w/o $\mathcal{L}_{gen}$).
    \item \textbf{V2.} This does not use the auxiliary training objective (w/o $\mathcal{L}_{aux}$). 
    \item \textbf{V3.} This skips mimicking the forward pass (Section~\ref{sec:forwardpass_mimicking}) in obtaining the flow surrogate (w/o Forward).
\end{itemize}

\vspace{-3mm}
\begin{itemize}[leftmargin=*]
    \item \textbf{V4.} This skips mimicking the backpropagation. (Section~\ref{subsec:backpropagation_mimicking}) in obtaining the flow surrogate (w/o Backward).
\end{itemize}
We use ResGatedGCN as our backbone encoder and evaluate under the performance prediction task.
Other settings are the same as in Section~\ref{subsec:performanceprediction}.

\textbf{Results.}
As shown in Table~\ref{tab:ablexp}, \method achieves superior performance over its four variants in five out of six settings, underscoring the necessity of its core components for achieving high performance.

\subsection{RQ5: Speed analysis}\label{subsec:speedanalysis}

We provide a speed analysis of \method.

\textbf{Setup.}
We analyze
(1) the total runtime and performance prediction results with (w/) and without (w/o) \method pre-training for the flow-based encoder (FlowerFormer) and the non-flow-based encoder (ResGatedGCN)
and (2) the pre-training runtime of each pre-training method, equipped with ResGatedGCN.
We set the batch size and pre-training epochs to 256 and 200, respectively.
Also, a comparison of all methods under equal runtime is in Appendix~\ref{subapp:fixedpretrainingtime}.





\begin{figure*}[t]
  \centering

  \begin{minipage}[t]{0.49\textwidth}
    \centering
    \includegraphics[width=\linewidth]{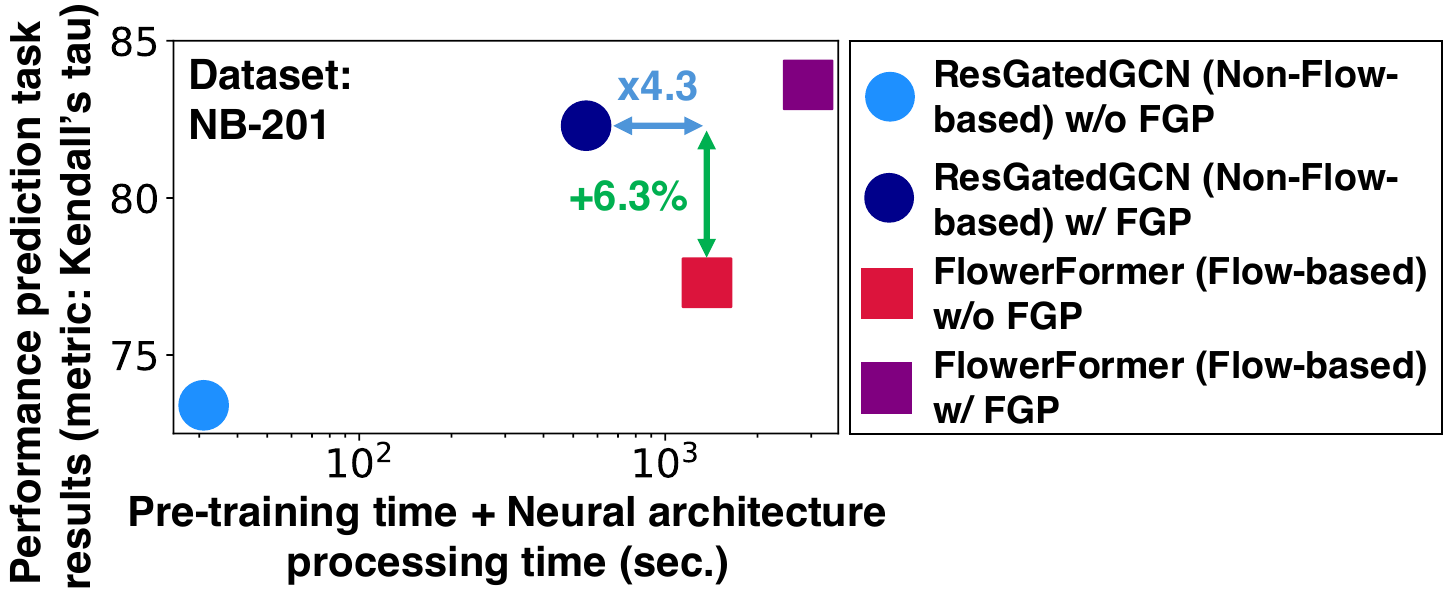}
    \small \textbf{(a)} Encoders’ runtime (pre-training time included) and performance-prediction results ($\times 100$).
  \end{minipage}
  \hfill
  \begin{minipage}[t]{0.49\textwidth}
    \centering
    \includegraphics[width=\linewidth]{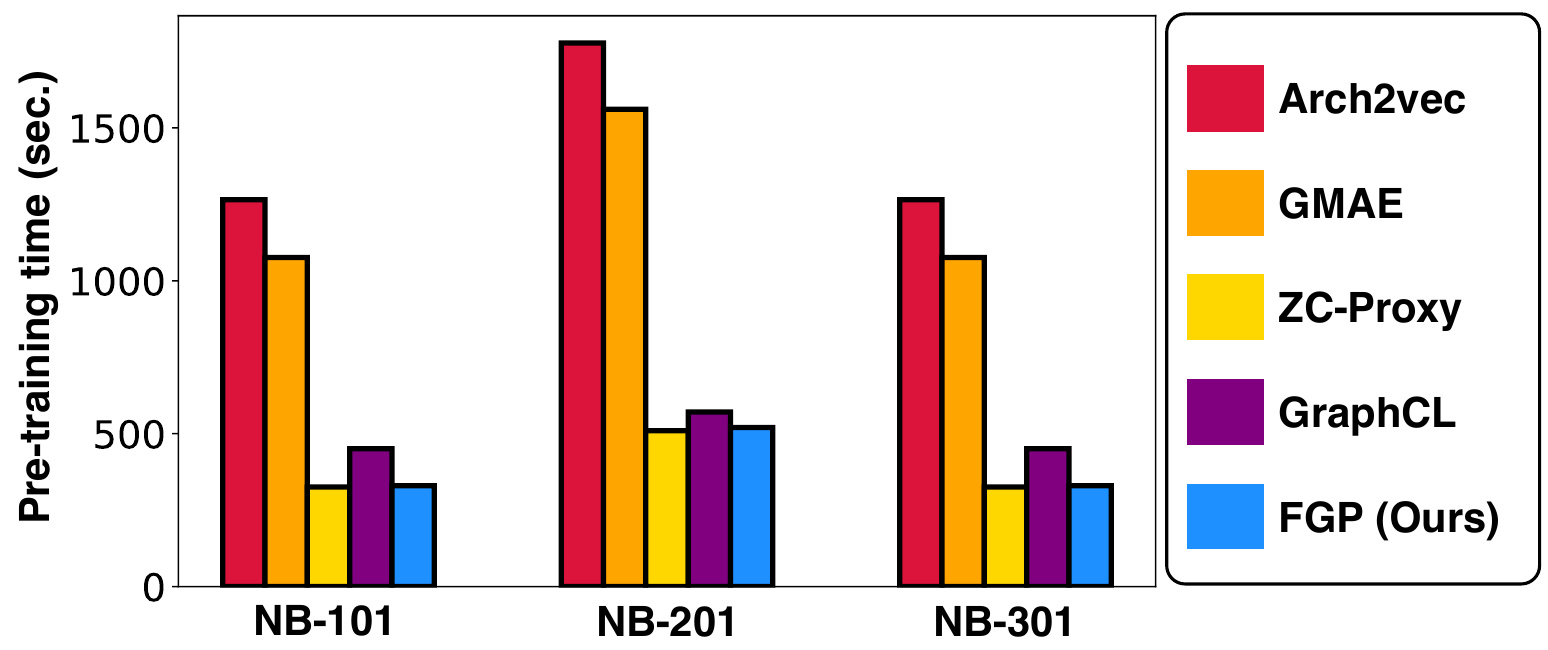}
    \small \textbf{(b)} Total pre-training time for each method using the ResGatedGCN backbone.
  \end{minipage}

  \caption{\textbf{Speed analysis.}
  (a) Pre-training the non-flow encoder (ResGatedGCN) with \method improves both performance and speed over the flow-based encoder (FlowerFormer) without \method.
  (b) \method is the second-fastest among the pre-training methods for neural architecture encoding.}
  \label{fig:speed_analysis}
  \vspace{3mm}
\end{figure*}

\textbf{Results.} 
First, as shown in Figure~\ref{fig:speed_analysis} (a),
pre-training the non-flow-based encoder (ResGatedGCN) with \method shows both performance and speed gain over the flow-based encoder (FlowerFormer) without \method in performance prediction. 
That is, despite this performance superiority, the total runtime for the pre-training and architecture processing of ResGatedGCN remains significantly shorter than the processing time alone of FlowerFormer.
Second, as shown in Figure~\ref{fig:speed_analysis} (b), \method is the second-fastest pre-training method, comparable to the fastest one, ZC-Proxy.
Thus, despite the large performance gain compared to the baselines, \method does not impose a significantly greater computational burden.

\section{Conclusion}
\label{sec:conclusion}
In this work, we propose a novel generative pre-training method for neural architecture encoding, called \method, which enables a neural architecture encoder to learn the information flow within a neural architecture, even without a specialized model structure.
To this end, \method trains a neural architecture encoder to reconstruct the flow surrogate, our proposed pre-training objective, based on information flow.
Our extensive experiments demonstrate the efficacy of \method in performance prediction and neural architecture search.
Code and datasets are in~\url{https://github.com/kswoo97/FGPAnom}.

\section*{Acknowledgements}
This work was supported by Institute of Information \& Communications Technology Planning \& Evaluation (IITP) grant funded by the Korea government (MSIT) (No. RS-2022-II220871, Development of AI Autonomy and Knowledge Enhancement for AI Agent Collaboration, 60\%)
(No. RS-2022-II220157, Robust, Fair, Extensible Data-Centric Continual Learning, 20\%)
(No. RS-2024-00457882, AI Research Hub Project, 10\%)
(RS-2019-II190075, Artificial Intelligence Graduate School Program (KAIST), 10\%).

\newpage

\bibliographystyle{abbrvnat}
\bibliography{neurips_2024}

\newpage 

\appendix

\section{Experimental details}
\label{app:expdetail}
\begin{table}[h]
\caption{\textbf{Dataset statistics.} 
Total counts of training and test data, along with the average time (in seconds) required to obtain the proposed flow surrogate (defined in Section~\ref{subsec:method_surrogate}).
}
\small
\centering
\scalebox{1.0}{
\renewcommand{\arraystretch}{1.1}
\begin{tabular}{l | c c c }
    \toprule

    Dataset & \# of train & \# of test & Time for surrogate (secs.) \\
    \midrule 
    \midrule 
     
     Nas-Bench-101~\citep{ying2019bench} & 7,290 & 7,290 & 0.002
     
     \\

     Nas-Bench-201~\citep{dong2020bench} & 7,813 & 7,812 & 0.002
    \\
    
     Nas-Bench-301~\citep{siems2020bench} & 5,611 & 5,610 & 0.005
     \\
     
     \bottomrule
\end{tabular}
\label{tab:datasetstat}
}
\vspace{5mm}
\end{table}

\subsection{Dataset details}\label{subapp:dataset}
In this section, we elaborate on the leveraged three convolution neural architecture datasets.
Dataset statistics are provided in Table~\ref{tab:datasetstat}.
Specifically, we represent each architecture based on an operation-on-node type graph~\citep{ning2022ta}, where each operation corresponds to a node, and each edge represents a connection between two operations.
\begin{itemize}[leftmargin=*]
    \item \textbf{NAS-Bench-101}~\citep{ying2019bench}. 
    This dataset includes 423K convolutional neural architectures, where each data point represents a specific neural architecture along with its test accuracy on the CIFAR-10 image classification task~\citep{krizhevsky2009learning}.
    We leverage a subset of the entire search space provided by~\citet{hwang2024flowerformer, ning2022ta}.
    \item \textbf{NAS-Bench-201}~\citep{dong2020bench}.
    This dataset includes 15K convolutional neural architectures, where each data point represents a specific neural architecture along with its test accuracy on the CIFAR-10 image classification task~\citep{krizhevsky2009learning}.
    \item \textbf{NAS-Bench-301}~\citep{siems2020bench}.
    This dataset includes $10^{16}$ convolutional neural architectures, where each data point represents a specific neural architecture along with its surrogate performance (i.e., an estimated performance produced by a performance predictor) on the CIFAR-10 image classification task~\citep{krizhevsky2009learning}.
    Each architecture is represented by two graphs: a normal cell and a reduction cell.
    Following~\citet{siems2020bench}, we connect the output node of the normal cell graph to the input nodes of the reduction cell, creating a single unified graph that represents the architecture.
    We leverage a subset of the entire search space provided by~\citet{abdelfattah2021zero}, where zero-cost proxies are publicly available.
\end{itemize}
For non-flow-based neural architecture encoders (i.e., ResGatedGCN~\citep{bresson2017residual} and GIN~\citep{xu2019powerful}), which lack specialized message passing for directed acyclic graphs, we transform each neural architecture graph into an undirected graph by adding a reverse edge for every existing graph.

\subsection{Machines and implementation}

We conducted our experiments on a machine with NVIDIA RTX 8000 D6 GPUs (48GB memory) and two Intel Xeon Silver 4214R processors.
\method is primarily implemented using the Pytorch (v1.12.1) and Pytorch Geometric (v2.2.0) libraries. 
We used AdamW~\citep{loshchilov2017decoupled} as the learning optimizer.

\subsection{Fine-tuning protocol}
In this section, we detail our fine-tuning protocol.
For a given neural architecture encoder $f_{\theta}$, which is randomly initialized or pre-trained with a particular pre-training method, we fine-tune the neural architecture encoder to perform the performance prediction task.

\noindent\textbf{Regression.}
Recall that a neural architecture encoder outputs a $d-$dimensional vector $\mathbf{z}^{(i)} \in \mathbb{R}^{d}$, representing a given neural architecture $\mathcal{G}^{(i)}$.
To estimate the performance of the corresponding neural architecture, denoted as ${y}^{(i)} \in \mathbb{R}$,
we employ a feed-forward-network-based regressor $r_{\xi} : \mathbb{R}^{d} \mapsto \mathbb{R}$.
The performance estimate $\hat{y}^{(i)}$ is obtained by projecting $\mathbf{z}^{(i)}$ using this regressor (i.e., $\hat{y}^{(i)} = r_{\xi}(\mathbf{z}^{(i)})$).

\noindent\textbf{Learning objective.}
Similar to our use of a zero-cost proxy (refer to Appendix~\ref{sec:zcusage}), to train a model to predict which architectures are `more likely' to perform well, we leverage a margin-ranking loss. 
For two neural architectures $\mathcal{G}^{(i)}$ and $\mathcal{G}^{(j)}$ within the same batch, we compute the following prediction loss:
\begin{equation}\label{eq:finetuningloss}
    \mathcal{L}_{fine} = \sum_{(i,j):y^{(i)}>y^{(j)}}\max{(0, m'-(\hat{y}^{(i)} - \hat{y}^{(j)}))},
\end{equation}
where $m'$ is a margin hyperparameter.
The parameters $\theta$ and $\xi$ are trained to minimize $\mathcal{L}_{fine}$ (Eq.~\ref{eq:finetuningloss}) via gradient descent.
This updating process continues until it satisfies the validation stopping criterion or reaches the maximum number of fine-tuning epochs.

\subsection{Hyperparameters}\label{app:hyperparameters}

\noindent\textbf{Neural architecture encoders.}
For ResGatedGCN~\citep{bresson2017residual} and GIN~\citep{xu2019powerful}, whose best hyperparameter configurations are not publicly provided, we tuned their hidden dimensions and maximum fine-tuning epochs within $\{64, 128, 256\}$ and $\{300,400,500\}$, according to the validation set performance on the performance prediction task. 
We fixed their number of layers, fine-tuning learning rate, and fine-tuning weight decay, as $3, 10^{-3},$ and $10^{-6}$, respectively.
After finding the hyperparameter configuration that gives the best validation performance, the chosen configuration is used across all the methods (i.e., the same encoder structure and pre-training scheme are leveraged across different pre-training methods).
For FlowerFormer~\citep{hwang2024flowerformer}, we follow the hyperparameter configurations provided in their official implementation~\url{https://github.com/y0ngjaenius/CVPR2024_FLOWERFormer}.

\noindent\textbf{Pre-training methods.}
For every pre-training method, including the baseline methods and our proposed method \method, we tuned the following hyperparameters:
\begin{itemize}[leftmargin=*]
    \item \textbf{Learning rate} within $\{10^{-3}, 5\times 10^{-4}, 10^{-4}\}$,
    \item \textbf{Projection head dimension} within $\{32, 64, 128,256\}$,
    \item \textbf{Projection head number of layers} within $\{1,2,3\}$,
    \item \textbf{Weight decay} within $\{10^{-6}, 0.0\}$.
\end{itemize}
Additionally, for \method, we tuned the learning objective coefficients $(\lambda_{1},\lambda_{2})$ within $\{(\frac{1}{2}, \frac{1}{2}), (\frac{1}{3}, \frac{2}{3}), (\frac{2}{3}, \frac{1}{3})\}$.

\subsection{Details regarding neural architecture search}\label{subapp:nasdetail}
In this section, we provide details regarding our neural architecture search experiment (Section~\ref{subsec:nas}).
Note that the process below is based on NPENAS~\cite{wei2022npenas}.

We first pre-train a neural architecture encoder using a designated pre-training strategy, such as \method. 
At the start of each round, the encoder is initialized with the pre-trained weights. 
The search begins with 20 neural architectures, each with known ground-truth performance. 
Using these, we fine-tune the encoder to predict performance. 
For each of the 20 architectures, we generate 5 valid candidates via mutation. 
Among the mutated architectures, we select the top 20 based on predicted performance and evaluate them. 
The selected architectures and their performances are added to the budget. 
With the updated pool, now containing 40 evaluated architectures, we repeat the process until the budget reaches a predefined size.

\section{Additional experimental results}
\label{app:additionalresult}
\begin{table*}[t]
\vspace{-2mm}
\caption{\textbf{Performance prediction results on various training set ratios.}
Mean and standard deviation on each dataset and metric.
The best performance is highlighted in \colorbox{sunwoogreen}{blue}.
All values are multiplied by 100.
N/A denotes an encoder solely trained with supervised learning, without any pre-training.
Gain from N/A denotes the performance improvement of \method compared to N/A.
\method outperforms the baselines in 14/18 cases.}
\setlength{\tabcolsep}{4.0pt}
\small
\centering
\scalebox{0.8}{
\renewcommand{\arraystretch}{0.9}
\begin{tabular}{c | c | c c c | c c c | c c c}
    \toprule
    \multirow{2}{*}{\makecell{Training \\ set ratio}} & \multirow{2}{*}{\makecell{Pre-training \\ method}} & \multicolumn{3}{c|}{Kendall's Tau} 
    & \multicolumn{3}{c|}{Precision@1\%} & \multicolumn{3}{c}{Precision@5\%} \\
    
    
         &  & NB-101 & NB-201 & NB-301
            & NB-101 & NB-201 & NB-301
            & NB-101 & NB-201 & NB-301 
            \\
    \midrule 
    \midrule 
    
    \multirow{7}{*}{5\%}
     
     & N/A & 75.8 {\std (3.4)} & 84.6 {\std (0.7)} & 67.2 {\std (1.1)}
     & 42.9 {\std (11.3)} & 54.3 {\std (3.7)} & 31.9 {\std (2.9)}
     & 64.5 {\std (3.6)} & 69.4 {\std (2.8)} & 51.9 {\std (3.4)}
     \\
     
     & GraphCL
     & 76.0 {\std (5.0)} & 85.4 {\std (0.7)} & 66.4 {\std (1.1)} 
     & {42.8 {\std (8.6)}} & 57.0 {\std (3.0)} & 29.9 {\std (3.8)} 
     & 63.4 {\std (6.3)} & 69.8 {\std (2.8)} & 50.6 {\std (1.9)} 
     \\
     
     & Arch2vec
     & 76.9 {\std (3.7)} & 85.3 {\std (0.7)} & 68.4 {\std (1.3)} 
     & 42.0 {\std (12.8)} & 56.7 {\std (5.0)} & 34.5 {\std (5.4)} 
     & 65.1 {\std (3.1)} & 70.2 {\std (3.1)} & 53.2 {\std (4.3)}
     \\

     & GMAE
     & 76.2 {\std (4.7)} & 87.0 {\std (0.5)} & 67.8 {\std (1.8)} 
     & 41.7 {\std (8.1)} & 59.0 {\std (5.4)} & \best 35.4 {\std (4.9)}
     & 62.5 {\std (5.6)} & 73.6 {\std (1.2)} & 53.1 {\std (3.5)} 
     \\

     & ZC-Proxy
     &  78.0 {\std (4.1)} & 87.7 {\std (0.4)} & 66.6 {\std (1.2)}
     &  42.7 {\std (7.5)} & 56.4 {\std (4.6)} & 33.9 {\std (4.4)}
     &  \best 67.3 {\std (3.0)} & 72.2 {\std (3.2)} & 51.7 {\std (4.7)}
     \\

     \cmidrule{2-11}

     & \method 
     & \best 79.4 {\std (3.3)} & \best 88.0 {\std (0.4)} &  \best 68.6 {\std (2.4)}  
     & \best 45.8 {\std (11.0)} & \best 59.7 {\std (5.4)} &  34.8 {\std (4.6)} 
     &  66.8 {\std (3.2)} & \best 73.6 {\std (2.2)} & \best 53.3 {\std (5.0)} 
     \\

     & Gain from N/A & +4.7\% & +4.0\% & +2.1\%
     & +6.8\% & +9.9\% & +9.1\% 
     & +3.6\% & +6.1\% & +2.7\% 
     \\
     
     \midrule 
     \midrule 

     \multirow{7}{*}{10\%}
     
     & N/A 
     & 77.3 {\std (5.1)} & 88.3 {\std (0.3)} & 69.9 {\std (1.2)} 
     & 43.2 {\std (8.9)} & 57.9 {\std (2.2)} & 35.4 {\std (5.2)} 
     & 65.8 {\std (4.0)} & 74.0 {\std (1.0)} & 56.6 {\std (2.0)}
     \\
     
     & GraphCL
     & 77.3 {\std (5.9)} & 88.9 {\std (0.3)} & 69.1 {\std (2.3)}
     & 42.7 {\std (9.0)} & 58.6 {\std (4.1)} & 35.6 {\std (3.7)}
     & 66.3 {\std (3.8)} & 73.3 {\std (1.3)} & 55.3 {\std (2.8)}
     \\
     
     & Arch2vec
     & 78.3 {\std (5.1)} & 88.8 {\std (0.4)} & 71.2 {\std (1.8)}
     & 43.2 {\std (11.6)} & 57.0 {\std (3.3)} & 37.2 {\std (5.6)}
     & 66.5 {\std (4.3)} & 73.9 {\std (1.1)} & 56.7 {\std (4.0)}
     \\

     & GMAE
     & 78.1 {\std (5.9)} &  89.9 {\std (0.4)} &   70.8 {\std (1.0)}
     &  45.5 {\std (10.6)} & \best 61.3 {\std (5.1)} & 38.8 {\std (4.8)}
     & 67.6 {\std (3.6)} & 75.4 {\std (0.7)} & \best 58.6 {\std (1.5)}
     \\

     & ZC-Proxy
     & 79.0 {\std (4.8)} & 89.7 {\std (0.2)} & 70.2 {\std (0.9)}
     & 44.0 {\std (9.0)} & 58.2 {\std (1.3)} & 37.4 {\std (4.4)}
     & 67.2 {\std (3.0)} & 74.5 {\std (1.4)} & 56.8 {\std (1.4)}
     \\

     \cmidrule{2-11}

     & \method 
     & \best 79.9 {\std (4.3)} & \best 89.9 {\std (0.3)} & \best 71.5 {\std (0.6)}
     & \best 45.8 {\std (10.2)} &  58.4 {\std (1.0)} & \best 38.9 {\std (5..0)}
     & \best 67.8 {\std (3.4)} & \best 76.0 {\std (1.0)} & 56.9 {\std (2.4)}
    \\

    & Gain from N/A & +3.4\% & +1.8\% & +2.3\%
     & +6.0\% & +1.0\% & +9.8\% 
     & +3.0\% & +2.7\% & +0.1\% 
     \\

     \bottomrule
\end{tabular}
\label{tab:addtrainexp}
}
\end{table*}

\subsection{Analysis of encoding time of diverse architecture encoders}\label{subapp:timecomparison}

In this section, we provide further details regarding the encoding time analysis, which is provided in Figure~\ref{fig:firstmotivation}.
Specifically, we provide detailed mean encoding time per architecture for the flow-aware neural architecture and the simple GNN-based neural architecture.

\noindent\textbf{Setup.}
We compare the encoding time of a flow-aware encoder (i.e., FlowerFormer) with non-flow-aware encoders (i.e., ResGatedGCN and GIN). Specifically, we measure the time required to generate representations for all neural architectures—both training and test—and report the average per architecture.

\begin{wraptable}{r}{0.5\textwidth}
\centering
\caption{Detailed mean encoding time per architecture of each neural architecture encoder}
\begin{tabular}{c|c|c}
\hline
& NB-101 & NB-201 \\
\hline
ResGatedGCN & 0.0021 & 0.0022  \\
GIN & 0.0013 & 0.0014 \\
FlowerFormer & 0.1242 & 0.0849 \\
\hline
\end{tabular}
\label{tab:detailencodertime}
\end{wraptable}

\noindent\textbf{Results.}
As shown in Table~\ref{tab:detailencodertime}, FlowerFormer, a flow-aware encoder, takes at most 94x more encoding time compared to GIN, which is a simple GNN encoder that does not have a specialized architecture to capture the information flow. 
This result supports our claim that flow-aware encoders are significantly slower than simple GNN-based encoders.

\subsection{Analysis under varying training set size}\label{subapp:additionalratio}

In this section, we present additional experimental results across different training set ratios. 
In the main paper, we focus on a scenario where only 1\% of the training set is used for fine-tuning, simulating a label-scarce scenario (i.e., a limited number of architecture-performance pairs available). 
Recall that this is because obtaining architecture-performance pairs involves significant computational costs, making it crucial to develop a pre-training method that enhances neural architecture encoders in label-scarce scenarios.
Furthermore, we evaluate each method in scenarios where additional architecture-performance pairs are available, corresponding to a larger training set ratio.

\noindent\textbf{Setup.}
We explore two additional settings for fine-tuning: using (1) 5\% and (2) 10\% of the training set, respectively.
We use ResGatedGCN~\citep{bresson2017residual} as the backbone neural architecture encoder.
Other experimental settings are the same as those used in our neural architecture performance prediction experiments (Section~\ref{subsec:performanceprediction}).

\noindent\textbf{Results.}
As shown in Table~\ref{tab:addtrainexp}, \method outperforms all baseline methods in 14 out of 18 settings.
This result demonstrates that the superiority of \method over existing pre-training methods is not restricted to a specific training set ratio setting.

\begin{table*}[t]
\vspace{-2mm}
\caption{\textbf{Performance prediction results when using only the training dataset even for pre-training.}
Mean and standard deviation on each dataset and metric.
The best performance is highlighted in \colorbox{sunwoogreen}{blue}.
All values are multiplied by 100.
N/A denotes an encoder solely trained with supervised learning, without any pre-training.
Gain from N/A denotes the performance improvement of \method compared to N/A.
\method outperforms the baseline methods in all the settings.}
\small
\centering
\scalebox{0.9}{
\renewcommand{\arraystretch}{1.0}
\begin{tabular}{c | c c  | c c | c c}
    \toprule
    \multirow{2}{*}{\makecell{Pre-training \\ method}} & \multicolumn{2}{c|}{Kendall's Tau} 
    & \multicolumn{2}{c|}{Precision@1\%} & \multicolumn{2}{c}{Precision@5\%} \\
    
    
         & NB-101 & NB-201 
            & NB-101 & NB-201 
            & NB-101 & NB-201 
            \\
    \midrule 
    \midrule 
     
     N/A 
     & 65.0 {\std (7.8)} & 73.4 {\std (0.7)}
     & 18.2 {\std (7.9)} & 29.7 {\std (3.7)} 
     & 46.2 {\std (12.2)} & 51.7 {\std (4.2)}
     \\
     
     GraphCL
     & 65.2 {\std (6.4)} & 74.3 {\std (1.6)} 
     & {20.1 {\std (12.6)}} & 36.4 {\std (9.7)} 
     & 44.4 {\std (11.3)} & 54.5 {\std (1.7)} 
     \\
     
     Arch2vec
     & 63.2 {\std (7.8)} & 73.5 {\std (1.9)} 
     & 18.5 {\std (9.3)} & 31.7 {\std (11.9)} 
     & 42.4 {\std (12.4)} & 53.3 {\std (5.7)} 
     \\

     GMAE
     & 67.3 {\std (5.2)} & 75.2 {\std (1.4)} 
     & 25.5 {\std (10.1)} & 34.8 {\std (9.9)}
     & 51.0 {\std (7.0)} & 56.8 {\std (4.9)} 
     \\

     ZC-Proxy
     &  69.8 {\std (5.1)} & 79.8 {\std (1.3)} 
     &  29.8 {\std (11.1)} & 39.2 {\std (12.8)}
     &  56.4 {\std (3.5)} & 58.9 {\std (8.2)} 
     \\

     \midrule

     \method 
     &  \best 73.5 {\std (5.5)} & \best 81.4 {\std (7.7)} 
     &  \best 37.0 {\std (10.9)} & \best 45.1 {\std (7.9)}
     &  \best 60.1 {\std (4.0)} & \best 61.5 {\std (2.8)}
     \\

     Gain from N/A & +13.1\% & +10.8\% 
     & +103.3\% & +51.9\% 
     & +30.1\% & +19.0\% 
     \\

     \bottomrule
\end{tabular}
\label{tab:onlytestresult}
}
\vspace{5mm}
\end{table*}

\subsection{Analysis under using only the training set, including for pre-training}
\label{appendix:strict}

In this section, we present additional experiment results when we only leverage the training dataset to perform pre-training.
In the main paper, we leverage the test dataset with the training during the pre-training, since architectures within the search space can be obtained with negligible cost. 
In the new setting, we aim to determine whether an encoder pre-trained with \method can generalize to unseen neural architectures.

\noindent\textbf{Setup.}
We do not use the test dataset for pre-training, and therefore, each test architecture becomes an unseen architecture.
We use ResGatedGCN~\citep{bresson2017residual} as the backbone neural architecture encoder.
Other experimental settings are the same as those used in our neural architecture performance prediction experiments (Section~\ref{subsec:performanceprediction}).

\noindent\textbf{Results.}
As shown in Table~\ref{tab:onlytestresult}, \method outperforms all baseline methods in all the settings.
This result demonstrates that the encoder trained with \method generalizes well to unseen neural architectures.

\begin{figure*}[t]
  \centering
{\includegraphics[width=\textwidth]{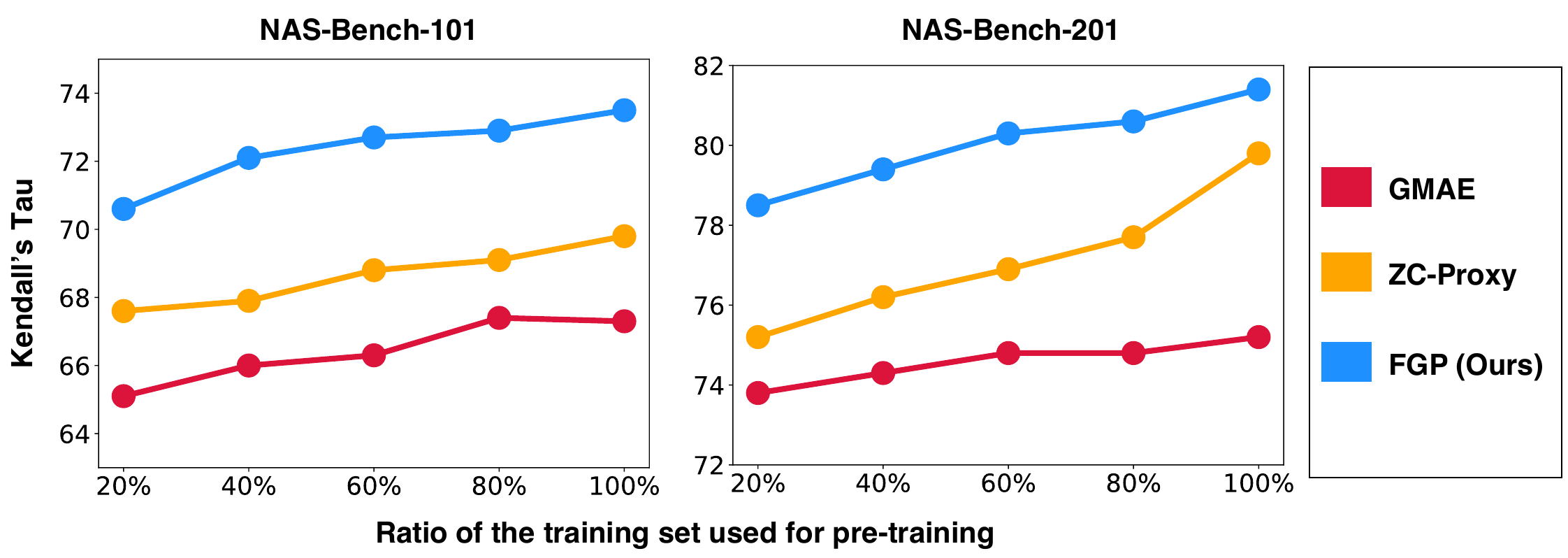}}
  \caption{\textbf{Performance under varying size of pre-training dataset.} 
  Performance prediction results when 20/40/60/80/100\% of the training set is used for pre-training of each method. 
  \method consistently outperforms the two strongest baseline methods (GMAE and ZC-Proxy) across all the settings.
  }
  \label{fig:differentsize}
\end{figure*}

\subsection{Analysis under varying pre-training set size}\label{subapp:differentpretrainingdatasize}

In this section, we present additional experimental results with varying pre-training dataset sizes, demonstrating that the effectiveness of \method is not restricted to a specific data size.

\noindent\textbf{Setup.} 
We randomly sample 20\%/40\%/60\%/80\%/100\% of the training set and use the sampled neural architectures for pre-training of a neural architecture, based on each pre-training strategy.
We use ResGatedGCN as the backbone neural architecture encoder, and the two strongest pre-training method (i.e., ZC-Proxy and GMAE) as baseline methods.
Other setups are the same as those used in our neural architecture performance prediction experiments (Section~\ref{subsec:performanceprediction}).

\noindent\textbf{Results.}
As shown in Figure~\ref{fig:differentsize}, the superiority of \method holds under varying size of pre-training dataset size, demonstrating that the effectiveness of \method is not limited to a particular dataset size.

\subsection{Analysis under a fixed pre-training time}\label{subapp:fixedpretrainingtime}

Recall that in Section~\ref{subsec:speedanalysis}, we provide analysis regarding the runtime of \method and other baseline methods.
In this section, we provide a performance comparison of pre-training methods under the fixed pre-trainig time.

\noindent\textbf{Setup.} 
We train a neural architecture for 100/200/300/400/500 seconds, by using each pre-training method.
When the corresponding time limit is reached, we execute the pre-training process and fine-tune and evaluate the pre-trained encoder for the performance prediction task.
We use ResGatedGCN as the backbone neural architecture encoder, and the two strongest pre-training method (i.e., ZC-Proxy and GMAE) as baseline methods.
Other setups are the same as those used in our neural architecture performance prediction experiments (Section~\ref{subsec:performanceprediction}).

\begin{figure*}[t]
  \centering
{\includegraphics[width=\textwidth]{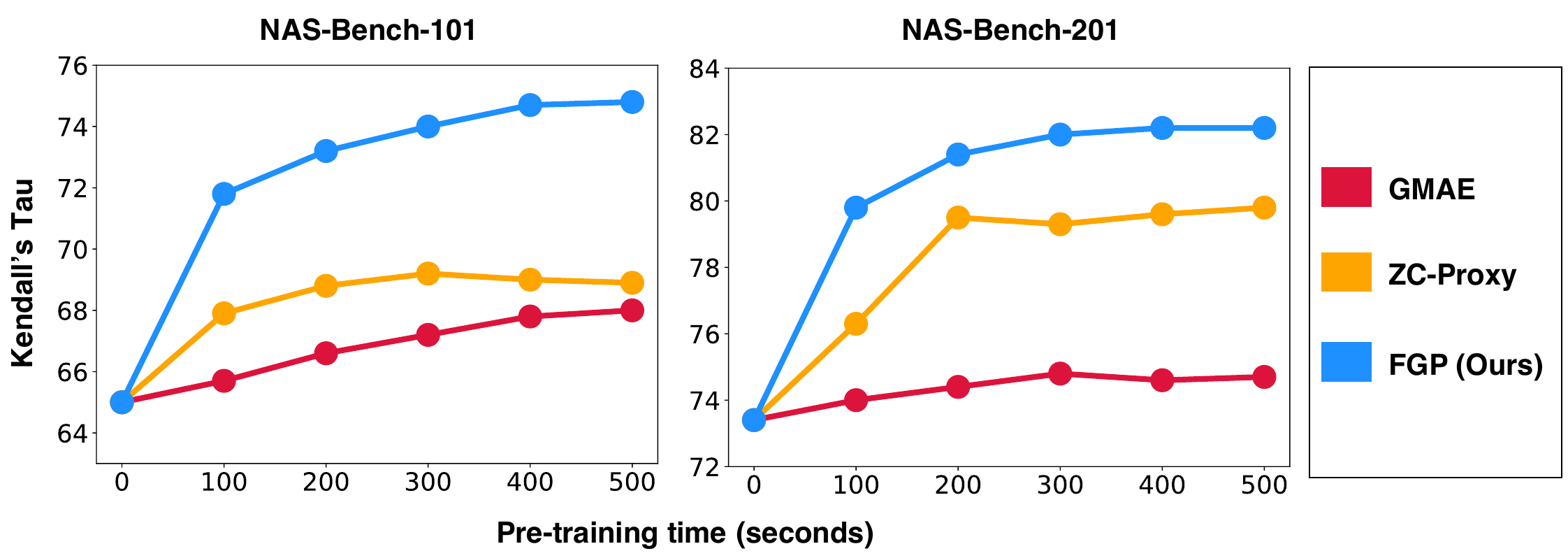}}
  \caption{\textbf{Performance under the fixed pre-training time.} 
  Performance prediction results under the fixed pre-training time of each pre-training method. 
  \method consistently outperforms the two strongest baseline methods (GMAE and ZC-Proxy) across all the settings.
  }
  \label{fig:fixedtime}
\end{figure*}

\noindent\textbf{Results.}
As shown in Figure~\ref{fig:fixedtime}, the superiority of \method holds under varying pre-training time split, demonstrating that the effectiveness of \method is not limited to a particular dataset size.

\subsection{Analysis under various domains}\label{subapp:newdomain}

In this section, we explore the effectiveness of \method in domains beyond computer vision, which are natural language processing and graph representation learning.

\begin{table*}[t]
\caption{\textbf{Performance prediction results under various domains.}
Mean and standard deviation on each dataset and metric.
The best performance is highlighted in \colorbox{sunwoogreen}{blue}.
All values are multiplied by 100.
Since zero-cost proxy values for the NB-Graph dataset are not publicly available, we could not run the ZC-Proxy baseline on the NB-Graph dataset, which is marked by `-'.
\method outperforms the baselines on natural language processing datasets and graph representation learning datasets.}
\small
\centering
\scalebox{1.0}{
\renewcommand{\arraystretch}{1.0}
\begin{tabular}{c | c c  | c c | c c}
    \toprule
    \multirow{2}{*}{\makecell{Pre-training \\ method}} & \multicolumn{2}{c|}{Kendall's Tau} 
    & \multicolumn{2}{c|}{Precision@1\%} & \multicolumn{2}{c}{Precision@5\%} \\
    
    
         & NB-NLP & NB-Graph
            & NB-NLP & NB-Graph
            & NB-NLP & NB-Graph
            \\
    \midrule 
    \midrule 
     
     N/A 
     & 28.5 {\std (6.1)} & 37.1 {\std (6.3)}
     & 2.9 {\std (1.4)} & 4.3 {\std (2.3)} 
     & 14.8 {\std (6.9)} & 17.1 {\std (4.1)}
     \\
     
     GraphCL
     & 28.7 {\std (5.3)} & 38.3 {\std (13.3)}
     & 3.1 {\std (1.3)} & 3.8 {\std (1.5)}
     & 14.9 {\std (6.1)} & 16.1 {\std (4.3)}
     \\
     
     Arch2vec
     & 28.4 {\std (5.7)} & 38.8 {\std (9.6)}
     & 3.0 {\std (1.2)} & 3.2 {\std (2.1)}
     & 15.1 {\std (5.2)} & 15.9 {\std (5.1)}
     \\

     GMAE
     & 28.9 {\std (6.6)} & 37.7 {\std (8.6)}
     & 2.2 {\std (1.0)} & 2.8 {\std (2.1)}
     & 15.7 {\std (5.3)} & 14.5 {\std (4.7)} 
     \\

     ZC-Proxy
     & 28.7 {\std (5.6)} & -
     & 2.9 {\std (2.3)} & -
     & 15.5 {\std (3.8)} & -
     \\

     \midrule

     \method 
     &  \best 29.7 {\std (4.8)} & \best 41.1 {\std (8.6)} 
     &  \best 4.3 {\std (2.5)} & \best 5.3 {\std (1.1)}
     &  \best 16.2 {\std (7.1)} & \best 18.3 {\std (3.5)}
     \\
     \bottomrule
\end{tabular}
\label{tab:otherdomain}
}
\end{table*}

\noindent\textbf{Setup.}
We use two additional benchmark datasets: (1) NAS-Bench-NLP~\citep{klyuchnikov2022bench}, a dataset that consists of natural language processing models, and (2) NAS-Bench-Graph~\citep{qin2022bench}, a dataset that consists of graph representation learning models.

\noindent\textbf{Results.}
As shown in Table~\ref{tab:otherdomain}, \method outperforms the baselines in 5 out of 6 settings, demonstrate that the effectiveness of \method does not limit to a computer vision domain.

\begin{table*}[t]
\caption{\textbf{Performance prediction results under a new downstream task.}
Mean and standard deviation on each dataset and metric.
N/A denotes a performance of an encoder without pre-training.
All values are multiplied by 100.
The best performance is highlighted in \colorbox{sunwoogreen}{blue}.
The superiority of \method over baseline methods holds valid under a new downstream task, dataset transfer.}
\small
\centering
\scalebox{1.}{
\renewcommand{\arraystretch}{1.0}
\begin{tabular}{l | c c  | c c }
    \toprule
    \multirow{2}{*}{\makecell{}} & \multicolumn{2}{c|}{Kendall's Tau} 
    & \multicolumn{2}{c}{Precision@10\%} \\
    
         &  TIN $\rightarrow$ DTD & TIN $\rightarrow$ OxfordPet 
            &  TIN $\rightarrow$ DTD & TIN $\rightarrow$ OxfordPet
            \\
    \midrule 
    \midrule 

     N/A
     & 62.7 {\std (4.6)} & 60.3 {\std (4.4)}
     & 50.0 {\std (31.1)} & 63.9 {\std (12.4)} \\

     GMAE
     & 63.4 {\std (3.8)} & 61.0 {\std (3.7)}
     & 55.5 {\std (25.8)} & 63.9 {\std (12.4)} \\

     ZC-Proxy
     & 64.1 {\std (4.9)} & 60.8 {\std (8.1)}
     & 47.0 {\std (29.7)} & 61.1 {\std (11.7)} \\

     \method
     & \best 65.3 {\std (4.3)} & \best 64.1 {\std (3.1)}
     & \best 63.9 {\std (12.4)} & \best 66.7 {\std (11.7)} \\

     \bottomrule
\end{tabular}
\label{tab:newtask}
}
\end{table*}

\subsection{Analysis under another task}\label{subapp:transferanalysis}

Recall that NAS-Bench-101, NAS-Bench-201, and NAS-Bench-301 are benchmark datasets containing image classification performances of various architectures. 
In this section, we assess each method's effectiveness in predicting architecture performance on a different task—dataset transferability—using a new dataset.

\noindent\textbf{Setup.}
Since no benchmark directly covers the transferability of architectures, we built a small neural architecture benchmark dataset.
To this end, we build two datasets: (1) TinyImageNet (TIN)~\cite{deng2009imagenet} $\rightarrow$ DTD~\cite{zhai2019large} and (2) TinyImageNet (TIN)~\cite{deng2009imagenet} $\rightarrow$ OxfordPet~\cite{zhai2019large}.
To this end, we randomly sampled 100 architectures from NAS-Bench-101, pre-trained each on TinyImageNet, and fine-tuned each on DTD or OxfordPet. 
We used each architecture’s test accuracy on the target dataset as the architecture’s transferability score.
Architectures were split 50/50 into train/test sets. 
ResGatedGCN is used as a backbone neural architecture encoder.
An architecture encoder was first pre-trained with a pre-training method (without transferability scores), then fine-tuned on the training set with transferability scores, and finally evaluated on the test set.

\noindent\textbf{Results.}
As shown in Table~\ref{tab:newtask}, \method outperforms all the baseline methods in all settings, demonstrating that its superiority extends beyond predicting image classification performance to also include architecture transferability prediction.
\subsection{Analysis under a new neural architecture encoder}\label{subapp:newencoder}

In this section, we explore the effectiveness of \method under a new directed-graph-based neural architecture encoder.

\noindent\textbf{Setup.} 
We use DiGCN~\cite{tong2020digraph}, a directed-graph-specialized neural architecture, as the backbone neural architecture encoder, and the two strongest pre-training methods (i.e., ZC-Proxy and GMAE) as baseline methods.
Other setups are the same as those used in our neural architecture performance prediction experiments (Section~\ref{subsec:performanceprediction}).

\noindent\textbf{Results.}
As shown in Table~\ref{tab:newencoder}, \method outperforms the baselines in all settings, demonstrate that the effectiveness of \method is not limited to a particular neural architecture encoder.

\begin{table*}[t]
\caption{\textbf{Performance prediction results under a new neural architecture encoder.}
Mean and standard deviation on each dataset and metric.
N/A denotes a performance of an encoder without pre-training.
All values are multiplied by 100.
The best performance is highlighted in \colorbox{sunwoogreen}{blue}.
The superiority of \method over baseline methods holds valid under a new encoder, DiGCN.}
\small
\centering
\scalebox{1.0}{
\renewcommand{\arraystretch}{1.0}
\begin{tabular}{l | c c  | c c | c c}
    \toprule
    \multirow{2}{*}{} & \multicolumn{2}{c|}{Kendall's Tau} 
    & \multicolumn{2}{c|}{Precision@1\%} & \multicolumn{2}{c}{Precision@5\%} \\
    
         & NB-101 & NB-201 
            & NB-101 & NB-201 
            & NB-101 & NB-201 
            \\
    \midrule 
    \midrule 

     N/A
     & 65.2 {\std (4.4)} & 65.7 {\std (1.5)}
     & 25.6 {\std (5.5)} & 14.1 {\std (4.8)} 
     & 49.0 {\std (1.1)} & 43.9 {\std (2.6)} \\

     GMAE
     & 65.1 {\std (3.9)} & 66.0 {\std (1.4)}
     & 26.3 {\std (5.4)} & 16.2 {\std (5.2)} 
     & 49.5 {\std (2.7)} & 44.2 {\std (2.1)} \\

     ZC-Proxy
     & 66.9 {\std (4.3)} & 72.1 {\std (1.3)}
     & 26.0 {\std (3.9)} & 18.8 {\std (4.7)} 
     & 50.5 {\std (5.7)} & 47.1 {\std (1.9)} \\

     \method
     & \best 73.3 {\std (3.7)} & \best 77.9 {\std (2.0)}
     & \best 37.3 {\std (8.6)} & \best 23.8 {\std (8.1)} 
     & \best 57.5 {\std (5.6)} & \best 50.9 {\std (5.8)} \\

     \bottomrule
\end{tabular}
\label{tab:newencoder}
}
\end{table*}

\noindent\textbf{Setup.}
We use two additional benchmark datasets: (1) NAS-Bench-NLP~\citep{klyuchnikov2022bench}, a dataset that consists of natural language processing models, and (2) NAS-Bench-Graph~\citep{qin2022bench}, a dataset that consists of graph representation learning models.

\noindent\textbf{Results.}
As shown in Table~\ref{tab:otherdomain}, \method outperforms the baselines in 5 out of 6 settings, demonstrate that the effectiveness of \method does not limit to a computer vision domain.

\subsection{Analysis regarding the usage of random vectors}\label{subsec:usingrandom}

In this section, we present additional experiment results regarding the usage of random vectors and matrices for our flow surrogate (Section~\ref{sec:method}).
We analyze two questions:
(Q1) Do the values of random vectors and matrices significantly impact the effectiveness of \method in downstream tasks?
(Q2) Does using random vectors and matrices for flow surrogates perform better than using representations of architectures obtained from flow-aware encoders?

\begin{figure*}[t]
  \centering
{\includegraphics[width=\textwidth]{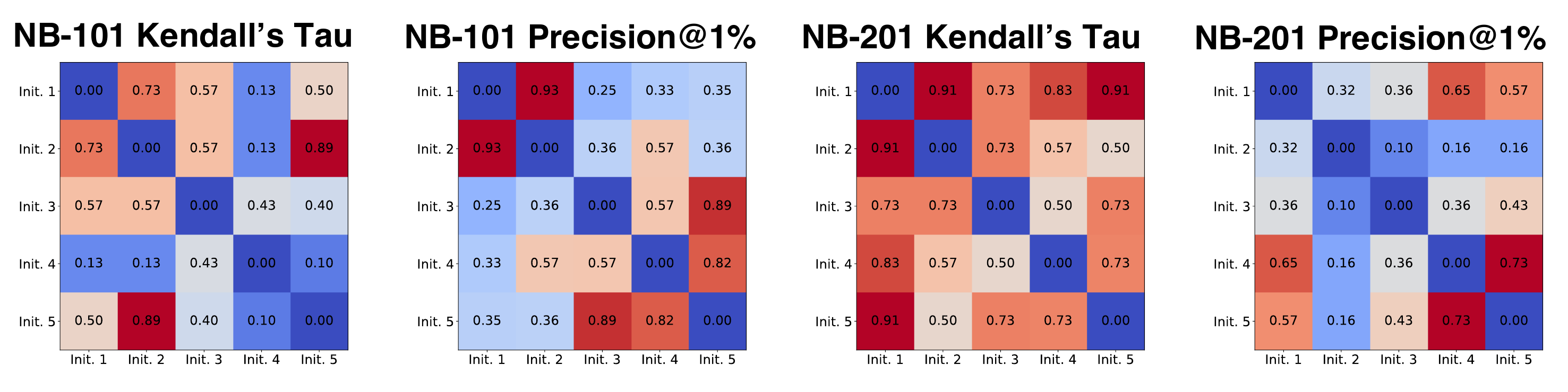}}
  \caption{\textbf{Statistical test across diverse random initialization for flow surrogates.} 
  p-values from the Wilcoxon Signed-Rank tests~\citep{wilcoxon1992individual} between pairs of performance sets, each obtained from different initializations, are reported. 
  At the significance level of $\alpha = 0.05$, none of the pairs provide sufficient statistical evidence to reject the null hypothesis. 
  Thus, different random initializations for the flow surrogate yield the same performance distribution.
  }
  \label{fig:randinit}
\end{figure*}

\begin{table*}[t]
\caption{\textbf{Performance prediction results under different derivations of flow surrogates.}
Mean and standard deviation on each dataset and metric.
All values are multiplied by 100.
The best performance is highlighted in \colorbox{sunwoogreen}{blue}.
Using random vectors to obtain flow surrogates outperforms using trained TA-GATEs as flow surrogates.}
\small
\centering
\scalebox{0.9}{
\renewcommand{\arraystretch}{1.0}
\begin{tabular}{l | c c  | c c | c c}
    \toprule
    \multirow{2}{*}{\makecell{Flow surrogates}} & \multicolumn{2}{c|}{Kendall's Tau} 
    & \multicolumn{2}{c|}{Precision@1\%} & \multicolumn{2}{c}{Precision@5\%} \\
    
         & NB-101 & NB-201 
            & NB-101 & NB-201 
            & NB-101 & NB-201 
            \\
    \midrule 
    \midrule 

    TA-GATEs~\citep{ning2022ta}
     & 72.2 {\std (5.6)} & 74.7 {\std (11.4)}
     & 29.2 {\std (7.2)} & 40.3 {\std (8.5)}
     & 54.4 {\std (7.3)} & 60.1 {\std (5.1)} \\
     
     Random vectors propagated as in Section~\ref{sec:method}
     & \best 74.8 {\std (4.8)} & \best 82.2 {\std (0.7)}
     & \best 37.5 {\std (13.0)} & \best 48.9 {\std (8.1)} 
     & \best 61.7 {\std (3.6)} & \best 62.3 {\std (2.5)} \\

     \bottomrule
\end{tabular}
\label{tab:varioustrial}
}
\end{table*}
\begin{table*}[t]
\caption{\textbf{Performance prediction results under variants of the flow surrogate.}
Mean and standard deviation on each dataset and metric.
All values are multiplied by 100.
The best performance is highlighted in \colorbox{sunwoogreen}{blue}.
Using only a single round flow surrogate yields the best performance.}
\small
\centering
\scalebox{1.0}{
\renewcommand{\arraystretch}{1.0}
\begin{tabular}{l | c c  | c c | c c}
    \toprule
    \multirow{2}{*}{\makecell{Propagation rounds}} & \multicolumn{2}{c|}{Kendall's Tau} 
    & \multicolumn{2}{c|}{Precision@1\%} & \multicolumn{2}{c}{Precision@5\%} \\
    
         & NB-101 & NB-201 
            & NB-101 & NB-201 
            & NB-101 & NB-201 
            \\
    \midrule 
    \midrule 
 
     Using only round 1 (Ours)
     & \best 74.8 {\std (4.8)} & \best 82.2 {\std (0.7)}
     & \best 37.5 {\std (13.0)} & \best 48.9 {\std (8.1)} 
     & \best 61.7 {\std (3.6)} &  62.3 {\std (2.5)} \\

     Using rounds 1 and 2
     & 71.7 {\std (5.4)} & 80.6 {\std (0.5)}
     & 36.7 {\std (5.9)} & 46.0 {\std (6.0)} 
     & 58.8 {\std (3.4)} & 62.0 {\std (1.9)} \\

     Using rounds 1, 2, and 3
     & 70.7 {\std (4.7)} & 81.6 {\std (0.8)}
     & 34.0 {\std (8.9)} & 47.3 {\std (0.6)} 
     & 58.2 {\std (4.1)} & \best 62.5 {\std (1.0)} \\

     \bottomrule
\end{tabular}
\label{tab:multiplerounds}
}
\vspace{2mm}
\end{table*}

\subsubsection{Q1. Various random initializations}

\noindent\textbf{Setup.}
We use five different random initializations of vectors and matrices to compute flow surrogates.
Then, we compare the performance of each initialization in the performance prediction task.
To this end, we use ResGatedGCN~\citep{bresson2017residual} as the backbone neural architecture encoder.
Other setups are the same as those used in our neural architecture performance prediction experiments (Section~\ref{subsec:performanceprediction}).

We leverage the Wilcoxon-Signed-Rank test~\citep{wilcoxon1992individual} to statistically compare the effects of different initialization. 
For each initialization setting, we conduct 9 performance prediction trials using flow surrogates generated with the corresponding initialization. 
For each pair of initializations, we apply the Wilcoxon Signed-Rank test to compare the performance sets derived from the respective initialization. 
The null hypothesis posits that the two performance distributions are identical, while the alternative hypothesis suggests they are distinct. 
We set the significance level at $\alpha = 0.05$.
If the p-value of the test exceeds $\alpha$, we lack sufficient statistical evidence to reject the null hypothesis, indicating that the performance sets from different initializations follow the same distribution.

\noindent\textbf{Results.}
As shown in Figure~\ref{fig:randinit}, every p-value is smaller than 0.05, resulting in we do not have sufficient statistical evidence to reject the null hypothesis.
These test results indicate that the performance sets derived from different random initializations of vectors and matrices for flow surrogates share the same distribution.

\subsubsection{Q2. Other choices for flow surrogates}

\noindent\textbf{Setup.}
For flow surrogates, we use architecture embeddings obtained by the flow-aware neural architecture encoder, TA-GATEs~\citep{ning2022ta}, trained through supervised learning.
In each experimental trial, we first train TA-GATEs using the training dataset.
Next, the trained TA-GATEs are employed to obtain neural architecture representations for every neural architecture in the search space.
Finally, these representations serve as flow surrogates for their corresponding neural architectures, which are then leveraged to pre-train another neural architecture encoder, ResGatedGCN with \method.

\noindent\textbf{Results.}
As shown in Table~\ref{tab:varioustrial}, flow surrogates derived using random vectors and matrices (as described in Section~\ref{sec:method}) outperform those derived using TA-GATEs, highlighting the effectiveness of flowing random vectors.
We hypothesize that this result stems from the potential overfitting of flow-aware encoders to the training dataset, causing them to focus on specific architectural patterns present in the training data.
In contrast, using random vectors likely avoids this bias, enabling the capture of broader patterns that are not limited to a specific subset of the dataset.

\subsection{Analysis regarding alternative design choices of \method}\label{subapp:newdesign}

Recall that the proposed flow surrogate is achieved after a single round of sequential message passing (refer to Section~\ref{subsec:method_surrogate}).
However, it is possible to perform message passing for multiple rounds.
In this section, we analyze the effectiveness of flow-surrogate variants, that is achieved via multiple rounds of sequential message passing.

\noindent\textbf{Setup.}
We use two variants:
\begin{itemize}[leftmargin=*]
    \item \textbf{(Variant 1)} This variant uses \textbf{two} surrogates: (1) a flow surrogate obtained via a single message passing round and (2) a flow surrogate obtained via two consecutive message passing rounds. It uses different projection heads for respective flow surrogate reconstruction. Reconstruction is performed in a joint manner.
    \item \textbf{(Variant 2)} This variant uses three surrogates: a flow surrogate obtained via (1) a single message passing round, (2) a flow surrogate obtained via two consecutive message passing rounds, and (3) a flow surrogate obtained via three consecutive message passing rounds. It uses different projection heads for respective flow surrogate reconstruction.
\end{itemize}
All methods use ResGatedGCN as a backbone neural architecture encoder.
Other setups are the same as those used in our neural architecture performance prediction experiments (Section~\ref{subsec:performanceprediction}).

\noindent\textbf{Results.}
As shown in Table~\ref{tab:multiplerounds}, a flow surrogate achieved via a single round of message passing, which is our proposed method, outperforms the alternatives in five out of six settings, justifying our design choice of flow surrogate.

    
 




\begin{table*}[t]
\caption{\textbf{Performance prediction results under various pooling functions.}
Mean and standard deviation on each dataset and metric.
All values are multiplied by 100.
The best performance is highlighted in \colorbox{sunwoogreen}{blue}.
Kendall's Tau is used as an evaluation metric.
Sum pooling outperforms other pooling functions.}
\small
\centering
\scalebox{1.0}{
\renewcommand{\arraystretch}{1.0}
\begin{tabular}{l | c  c c}
    \toprule
     & Mean (variant 1) & Max (variant 2) & Sum (Ours) \\

    \midrule

    NB-101 & 71.6 {\std (5.8)} & 71.1 {\std (6.4)} & 74.8 {\std (4.8)} \best \\
    NB-201 & 81.7 {\std (1.0)} & 81.5 {\std (0.9)} & 82.2 {\std (0.7)} \best \\

     \bottomrule
\end{tabular}
\label{tab:pooling}
}
\end{table*}

\subsection{Alternative pooling strategies}\label{subapp:pooling}
Note that we use the sum pooling to aggregate the messages from the neighbors (Section~\ref{subsec:method_surrogate}).
In this section, we analyze the alternative pooling functions that are widely used in GNN literature.

\textbf{Setup.}
We use the two strategies: (1) mean pooling and (2) max pooling.
Specifically, we replace the sum pooling function with one of the two functions and use the resulting flow surrogate as the final flow surrogate.
We use ResGatedGCN as the backbone neural architecture encoder, and other settings are the same as in Section~\ref{subsec:performanceprediction}.

\textbf{Results.}
As shown in Table~\ref{tab:pooling}, sum pooling, which \method uses to obtain flow surrogate, outperforms alternative strategies, demonstrating the effectiveness of our design choice for pooling function.

    
 




\begin{table*}[t]
\caption{\textbf{Operation classification results.}
Mean and standard deviation on each dataset and metric.
All values are multiplied by 100.
In every case, flow surrogates achieve accuracy higher than 92.}
\small
\centering
\scalebox{1.0}{
\renewcommand{\arraystretch}{1.0}
\begin{tabular}{l | c  c c}
    \toprule
    & $1\times 1$ Conv.  & $3\times3$ Conv. & Pooling \\

    \midrule
    \midrule

    NB-101 & 93.2 {\std (0.3)} & 99.9 {\std (0.1)} & 92.1 {\std (0.7)} \\ 

    NB-201 & 98.9 {\std (0.3)} & 99.9 {\std (0.1)} & 1.0 {\std (0.1)} \\

     \bottomrule
\end{tabular}
\label{tab:conv}
}
\end{table*}

\subsection{Expressiveness of our flow surrogate}\label{subapp:expressiveness}
In this section, we provide preliminary analysis regarding the expressiveness of our flow surrogate. 

\textbf{Setup.}
We analyze whether our flow surrogate can discriminate architectures containing different operations.
We conduct three binary classification tasks to determine whether a given neural architecture includes: (1) a 1×1 convolution operation, (2) a 3×3 convolution operation, and (3) a pooling operation. 
We use flow surrogate representations as input features and train an MLP as the classifier, performing 10 trials. The architectures are split into 80\% for training and 20\% for testing.

\textbf{Results.}
As shown in Table~\ref{tab:conv}, our flow surrogate achieves higher than 92\% accuracy in all the cases.
This result suggests that our flow surrogate can effectively distinguish architectures containing different operations, achieving

\subsection{Theoretical properties of our flow surrogate}\label{subapp:theory}
In this section, we investigate the theoretical properties of our flow surrogate.
In a nutshell, we provide a preliminary theoretical analysis demonstrating that our flow-surrogate computation is invariant to node permutations (i.e., node indexing), a desirable property when representing the information flow of a neural architecture to ensure effective downstream task performance.

\textbf{Motivation.}
\cite{hwang2024flowerformer, ning2022ta} demonstrates the importance of accurately capturing information flow for the downstream tasks (e.g., performance prediction and neural architecture search).
In the NAS-Bench datasets used in our experiments, the information flow within a neural architecture is independent of the indexing of individual operations. 
Accordingly, it is desirable for the computation of flow surrogates to be \textit{permutation-invariant} with respect to these operations.

\textbf{Proof sketch}
\citet{thost2021directed} theoretically demonstrates that asynchronous message passing on a directed acyclic graph—also used in our flow-surrogate computation—is invariant to node permutations when the permutation-invariant pooling function is employed for neighbor aggregation. 
Their theoretical result extends to our setting, indicating that our method likewise ensures permutation invariance regarding node indexing.

\begin{table*}[t]
\caption{\textbf{Performance under diverse backbone architecture encoders.}
Mean and standard deviation on each dataset and metric.
All values are multiplied by 100.
Kendall's Tau is used as an evaluation metric.
The best performance is highlighted in \colorbox{sunwoogreen}{blue}.
Since the MLP-based model does not return node embeddings, GMAE is not applicable.
Across diverse backbone encoders, \method consistently outperforms the baseline methods.}
\small
\centering
\scalebox{1.0}{
\renewcommand{\arraystretch}{1.0}
\begin{tabular}{l l | c c  c c}
    \toprule
    Backbone & Datasets& w/o pre-training & GMAE & ZC-Proxy & \method \\

    \midrule
    \midrule

    \multirow{2}{*}{MLP-based} & NB-101 & 42.5 {\std (5.4)} & - & 49.8 {\std (7.3)} & \best 65.2 {\std (4.1)} \\
    & NB-201 & 59.6 {\std (2.2)} & - & 70.2 {\std (1.8)} & 72.3 {\std (1.3)} \best \\

    \multirow{2}{*}{Transformer-based} & NB-101 & 62.7 {\std (3.6)} & 63.2 {\std (4.5)} & 64.1 {\std (8.5)} & 69.7 {\std (6.3)} \best \\
    & NB-201 & 63.2 {\std (2.0)} & 65.7 {\std (2.4)} & 69.2 {\std (3.2)} & 74.3 {\std (1.5)} \best \\

     \bottomrule
\end{tabular}
\label{tab:manybackbone}
}
\end{table*}

\subsection{Alternative backbone neural architecture encoders}\label{subapp:backbone}

Recall that we use graph-based neural architecture encoders for our experiments, given that neural architectures are often expressed as graphs.
In this section, we investigate the effectiveness of \method under non-graph-based neural architectures.

\textbf{Setup.}
We use two non-graph-based neural architecture encoders: MLP-based model~\cite{white2020study} and Transformer-based model~\cite{lu2021tnasp}.
Specifically, we first pre-train each architecture encoder via a certain pre-training method, and then fine-tune the pre-trained encoder with the performance prediction task.
Other settings are the same as in Section~\ref{subsec:performanceprediction}.

\textbf{Results.}
As shown in Table~\ref{tab:manybackbone}, \method consistently outperforms the baseline methods across diverse non-graph-based backbone neural architecture encoders, demonstrating that the effectiveness of \method is not limited to a particular graph-based backbone encoder.

    
 




\begin{table*}[t]
\caption{\textbf{Hyperparameter sensitivity analysis.}
Mean and standard deviation on each dataset and metric.
All values are multiplied by 100.
Kendall's Tau is used as an evaluation metric.
Across diverse settings of $\lambda_1$ and $\lambda_2$, \method consistently outperforms the baseline methods.}
\small
\centering
\scalebox{1.0}{
\renewcommand{\arraystretch}{1.0}
\begin{tabular}{l | c  c c c c}
    \toprule
    & $(\lambda_1,\lambda_2)=(\frac{1}{2},\frac{1}{2})$  & $(\lambda_1,\lambda_2)=(\frac{1}{3},\frac{2}{3})$ &
    $(\lambda_1,\lambda_2)=(\frac{2}{3},\frac{1}{3})$ &  ZC-Proxy & GMAE \\

    \midrule
    \midrule

    NB-101 & 74.8 {\std (4.8)} & 74.8 {\std (4.4)} & 74.3 {\std (4.5)} & 68.3 {\std (6.7)} & 68.1 {\std (4.7)} \\ 

    NB-201 & 82.2 {\std (0.7)} & 82.3 {\std (0.8)} & 81.7 {\std (0.8)} & 79.9 {\std (0.8)} & 74.8 {\std (1.2)} \\

     \bottomrule
\end{tabular}
\label{tab:hyperparameter}
}
\end{table*}

\subsection{Hyperparameter sensitivity}\label{subapp:hyperparameter}

In this section, we investigate the hyperparameter sensitivity of \method. 

\textbf{Setup.}
We analyze the coefficients of each loss term, which are $\lambda_{1}$ and $\lambda_{2}$ described in Section~\ref{subsec:method_pipeline}.
As detailed in Appendix~\ref{app:hyperparameters}, search space of the coefficients are as follows:
$(\lambda_1, \lambda_2) \in \{(\frac{1}{3}, \frac{2}{3}), (\frac{1}{2}, \frac{1}{2}), (\frac{2}{3}, \frac{1}{3})\}$.
Therefore, we measure the performance of \method under each configuration. 
We use ResGatedGCN as the backbone neural architecture encoder, and other settings are the same as in Section~\ref{subsec:performanceprediction}.

\textbf{Results.}
As shown in Table~\ref{tab:hyperparameter}, across diverse settings of $\lambda_1$ and $\lambda_2$, \method consistently outperforms the baseline methods, demonstrating the robustness of \method under the choice of $\lambda_1$ and $\lambda_2$.

    
 




\begin{table*}[t]
\caption{\textbf{FlowerFormer analysis under diverse pre-training data size.}
Mean and standard deviation on each dataset and metric.
All values are multiplied by 100.
Kendall's Tau is used as an evaluation metric.
Performance of the FlowerFormer pre-trained with \method tends to increase as the size of the pre-training data increases.}
\small
\centering
\scalebox{1.0}{
\renewcommand{\arraystretch}{1.0}
\begin{tabular}{l | c  c c c c c }
    \toprule
    Pre-training data ratio & 0\% & 20\% & 40\% & 60\% & 80\% & 100\%  \\

    \midrule
    \midrule

    NB-101 & 74.0 {\std (3.6)} & 74.6 {\std (3.8)} & 75.9 {\std (1.9)} & 75.8 {\std (4.0)} & 76.2 {\std (5.1)} & 76.3 {\std (3.6)} \\ 

    NB-201 & 77.3 {\std (1.5)} & 81.1 {\std (0.8)} & 82.3 {\std (1.2)} & 82.4 {\std (1.3)} & 82.7 {\std (1.4)} & 83.5 {\std (1.7)} \\

     \bottomrule
\end{tabular}
\label{tab:flowerformer}
}
\vspace{5mm}
\end{table*}

\subsection{Analysis regarding performance gain in FlowerFormer}\label{subapp:flowerformergain}

Recall that FlowerFormer~\cite{hwang2024flowerformer} is a flow-based encoder, which captures information flow within the neural architecture by using its architectural design.
However, as shown in Table~\ref{tab:mainexp}, the performance of FlowerFormer even improves with \method. 
In this section, we further investigate this phenomenon.
To this end, we first present the high-level hypothesis regarding the reason behind this phenomenon, and provide experiments that further support our hypothesis.

\textbf{Hypothesis.}
We hypothesize that FGP enables FlowerFormer to learn a broader range of information flow beyond what is present in labeled architectures by leveraging a large number of unlabeled architectures, thereby enhancing its ability to capture more diverse flow patterns.
When trained solely on labeled samples, FlowerFormer tends to learn information flow patterns limited to that specific set. In contrast, FGP leverages a much larger pool of neural architectures whose ground-truth performance is unknown, exposing FlowerFormer to a wider range of information flow patterns. This exposure enables FlowerFormer to learn more diverse and generalizable information flow representations.

\textbf{Setup.}
By varying the number of neural architectures used during FGP pre-training, we aim—though not precisely—to control the diversity of information flow patterns to which FlowerFormer is exposed. 
A positive correlation between dataset size and FlowerFormer performance suggests that FGP helps FlowerFormer encounter a broader range of information flow patterns, leading to improved accuracy in performance prediction.
We vary the proportion of the pre-training dataset used for FGP training—0\% (no pre-training), 20\%, 40\%, 60\%, 80\%, and 100\%—and evaluate FlowerFormer’s performance on the performance prediction task. 
Kendall’s Tau is used as the evaluation metric, and all other settings follow those described in Section~\ref{subsec:performanceprediction}.

\textbf{Results.}
As shown in Table~\ref{tab:flowerformer}, the performance of FlowerFormer pre-trained with \method tends to increase as the size of the pre-training dataset increases. 
This result supports our hypothesis on the performance improvement of the flow-based encoder (FlowerFormer) with \method.

\section{Flow surrogate details}
\label{app:surrogate}
\subsection{Details of node embeddings and messages}
In this section, we elaborate on the node embeddings (i.e., operation representations) $\mathbf{h}_{i} \in \mathbb{R}^{k}, \forall v_{i} \in \mathcal{V}$, messages of the order-1 nodes $\mathbf{r}$,~\footnote{Note that all the order-1 nodes share the same embedding $\mathbf{r}$.} and projection matrix $\mathbf{W} \in \mathbb{R}^{2k \times k}$.
\begin{itemize}[leftmargin=*]
    \item \textbf{Node embeddings.} We first randomly generate a transformation matrix $\mathbf{P} \in \mathbb{R}^{\vert \mathcal{O}\vert \times k}$, where $\mathcal{O}$ is a set of all the possible operations.
    Each entry of $\mathbf{P}$ is sampled from $\mathcal{N}(0, \sigma^{2})$ independently to each other.
    Then, we multiply $\mathbf{P}$ with node features $\mathbf{X}$, obtaining the node embedding matrix $\mathbf{H}$ (i.e., $\mathbf{H} = \mathbf{X}\mathbf{P}$), where $\mathbf{H}_{i,:} = \mathbf{h}_{i}$ holds.
    \item \textbf{Messages of the order-1 nodes.}
    We randomly sample each element of $\mathbf{r}$ from the uniform distribution $U(0, 1)$.
    \item \textbf{Projection matrix.} We randomly sample each element of $\mathbf{W}$ from $\mathcal{N}(0, \sigma^{2})$.
\end{itemize}

\subsection{Hyperparameters for obtaining flow surrogate}
We provide a search space or a fixed value for each hyperparameter related to the flow surrogate:
\begin{itemize}[leftmargin=*]
    \item \textbf{Standard deviation of Gaussian distribution $\sigma$.} This is tuned within $\{10^{-1}, 10^{0}\}$.
    \item \textbf{Dimension $k$.} This is tuned within $\{4, 5, \cdots , 12\}$.
    \item \textbf{Identity coefficient $\alpha$.} This is fixed as $\alpha = 0.5$.
\end{itemize}






\section{Zero-cost proxy details}
\label{app:zerocostproxy}
In this section, we provide further details regarding zero-cost proxy, which we use its prediction as an auxiliary training objective of \method.

\subsection{Descriptions of zero-cost proxies}
\noindent\textbf{Overview.} 
Zero-cost proxies are pruning-at-initialization metrics that represent certain characteristics of a neural architecture~\citep{abdelfattah2021zero}.
For a given neural architecture, these metrics can be obtained by simply building a deep learning model having the corresponding neural architecture (e.g., $\ell_{2}-$norm of parameters) or performing a single forward pass and gradient computation (e.g., $\ell_{2}-$norm of parameters' gradients).
Therefore, compared to actual model training—which requires a substantial number of training iterations (i.e., forward passes and backpropagation)—obtaining zero-cost proxies is significantly more economical~\citep{abdelfattah2021zero}.

\noindent\textbf{Usage of zero-cost proxies.}
Numerous studies on Neural Architecture Search (NAS) have demonstrated that these metrics exhibit a strong correlation with the ground-truth performance of neural architectures~\citep{abdelfattah2021zero, zhao2023dynamic, huang2024up}.
Consequently, various NAS approaches leverage zero-cost proxies to quickly assess the potential effectiveness of candidate architectures for the target task and dataset~\citep{huang2024up, lee2024az}.
Similarly, \citet{zhao2023dynamic} proposed pre-training a neural architecture encoder to predict the zero-cost proxies of a neural architecture, enabling the encoder to identify architectures more likely to achieve high performance.
Note that this zero-cost-prediction-based pre-training~\citep{zhao2023dynamic} has been leveraged as our baseline method, which is called ZC-Proxy.
Specifically, ZC-Proxy pre-trains the encoder following the scheme described in Appendix~\ref{sec:zcusage}. 

\subsection{Our usage of zero-cost proxy}\label{sec:zcusage}
\noindent\textbf{Overview.}
Motivated by the recent success of zero-cost proxies in NAS, we use predicting a zero-cost proxy of a neural architecture as an auxiliary learning objective $\mathcal{L}_{aux}$.
We use Jacobian Covariance zero-cost proxy~\citep{abdelfattah2021zero}, measuring how well a neural architecture distinguishes distinct inputs.
We train a neural architecture encoder to predict the Jacobian Covariance of a given neural architecture.
For NAS-Bench-101~\citep{ying2019bench} and NAS-Bench-301~\citep{siems2020bench}, we use proxies provided by~\citet{krishnakumar2022bench}.
For NAS-Bench-201~\citep{dong2020bench}, we use proxies provided by~\citet{abdelfattah2021zero}.

\noindent\textbf{Usage.}
Consider $\mathbf{z}^{(i)} \in \mathbb{R}^{d}$, an embedding of a neural architecture $\mathcal{G}^{(i)}$, obtained by a neural architecture encoder $f_{\theta}$.
We use a zero-cost-proxy regressor $u_{\rho}$ to predict the architecture's zero-cost proxy, denoted as $c^{(i)} \in \mathbb{R}$.
Specifically, the predicted zero-cost proxy $\hat{c}^{(i)} \in \mathbb{R}$ is obtained as follows: $\hat{c}^{(i)} = u_{\rho}(\mathbf{z}^{(i)})$.
Since the objective of learning zero-cost proxies is to train a model to identify which architectures are \textit{more likely} to perform better, we adopt a ranking-based loss. 
This loss encourages the model to focus on \textit{relative performance}, specifically identifying which architecture outperforms another.
To achieve this, we use margin ranking loss, which is widely used in training a neural architecture encoder~\citep{ning2022ta, hwang2024flowerformer}.
The loss function is formalized as:
\begin{equation}\label{eq:marginranking}
    \mathcal{L}_{aux} = \sum_{(i,j):c^{(i)}>c^{(j)}}\max{(0, m-(\hat{c}^{(i)} - \hat{c}^{(j)}))},
\end{equation}
where $\mathcal{G}^{(i)}$ and $\mathcal{G}^{(j)}$ are two neural architectures within the same batch, and $m$ is a margin hyperparameter.
The parameters $\theta$ and $\rho$ are optimized using gradient descent to minimize the auxiliary loss $\mathcal{L}_{aux}$ (Eq.~\ref{eq:marginranking}).

\section{Discussions}
\label{app:limitations}

In this section, we discuss limitations and broader impacts of our research.

\subsection{Limitations and potential future works}

While our work demonstrates empirical effectiveness in neural architecture encoding across various benchmark datasets and applications (Section~\ref{sec:experiment}), the theoretical properties of \method remain underexplored—specifically, what types of flows it can or cannot capture. 
While we provide preliminary analysis regarding these concepts in Appendices~\ref{subapp:expressiveness}~and~\ref{subapp:theory}, a further rigorous and in-depth analysis regarding them would enrich our understanding of our method.
Thus, further theoretical investigation into our flow surrogate offers promising directions for improvement.
In addition, given findings that LLMs understand graphs and produce interpretable graph representations~\cite{kim2025hello}, incorporating them into \method\ is a promising path to improve interpretability.
Moreover, several recent methods represent neural architectures in diverse formats, such as hypergraph~\citep{lin2024hypergraph}, which model interactions among multiple nodes~\citep{kim2024survey}.
Therefore, extending \method to such approaches would be an interesting direction.

\subsection{Broader impacts}

While our work focuses on neural architecture encoding, our approach has the potential to generalize to other data structures that exhibit a notion of \textit{flow}, such as electrical circuits~\cite{bakshi2020electrical} and Petri Nets~\cite{peterson1977petri}.
Accordingly, our flow surrogate presents practical opportunities for representing such structures in these domains.

\newpage

\section{NeurIPS Paper Checklist}

\begin{enumerate}[leftmargin=*]

\item {\bf Claims}
    \item[] Question: Do the main claims made in the abstract and introduction accurately reflect the paper's contributions and scope?
    \item[] Answer: \answerYes{} 
    \item[] Justification: We propose a generative-pretraining method for neural architecture encoding.

\item {\bf Limitations}
    \item[] Question: Does the paper discuss the limitations of the work performed by the authors?
    \item[] Answer: \answerYes{} 
    \item[] Justification: We discuss them in Appendix~\ref{app:limitations}.
    \item[] Guidelines:

\item {\bf Theory assumptions and proofs}
    \item[] Question: For each theoretical result, does the paper provide the full set of assumptions and a complete (and correct) proof?
    \item[] Answer: \answerYes{} 
    \item[] Justification: We discuss them in Appendix~\ref{subapp:theory}.

    \item {\bf Experimental result reproducibility}
    \item[] Question: Does the paper fully disclose all the information needed to reproduce the main experimental results of the paper to the extent that it affects the main claims and/or conclusions of the paper (regardless of whether the code and data are provided or not)?
    \item[] Answer: \answerYes{} 
    \item[] Justification: We present detailed experimental settings and hyperparameters, together with the source code.

\item {\bf Open access to data and code}
    \item[] Question: Does the paper provide open access to the data and code, with sufficient instructions to faithfully reproduce the main experimental results, as described in supplemental material?
    \item[] Answer: \answerYes{} 
    \item[] Justification: We make all our code and datasets publicly available through \url{https://github.com/kswoo97/FGPAnom}.

\item {\bf Experimental setting/details}
    \item[] Question: Does the paper specify all the training and test details (e.g., data splits, hyperparameters, how they were chosen, type of optimizer, etc.) necessary to understand the results?
    \item[] Answer: \answerYes{} 
    \item[] Justification: We provide detailed settings and hyperparameter search space of our experiments.

\item {\bf Experiment statistical significance}
    \item[] Question: Does the paper report error bars suitably and correctly defined or other appropriate information about the statistical significance of the experiments?
    \item[] Answer: \answerYes{} 
    \item[] Justification: We provide mean and standard deviation of our experimental results obtained via multiple trials.

\item {\bf Experiments compute resources}
    \item[] Question: For each experiment, does the paper provide sufficient information on the computer resources (type of compute workers, memory, time of execution) needed to reproduce the experiments?
    \item[] Answer: \answerYes{} 
    \item[] Justification: We provide detailed information regarding the machines used in our experiments.
    
\item {\bf Code of ethics}
    \item[] Question: Does the research conducted in the paper conform, in every respect, with the NeurIPS Code of Ethics \url{https://neurips.cc/public/EthicsGuidelines}?
    \item[] Answer: \answerYes{} 
    \item[] Justification: We carefully checked the NeurIPS Code of Ethics, and believe that we did not violate any of the ethical terms.

\item {\bf Broader impacts}
    \item[] Question: Does the paper discuss both potential positive societal impacts and negative societal impacts of the work performed?
    \item[] Answer: \answerYes{} 
    \item[] Justification: We discuss them in Appendix~\ref{app:limitations}.

\item {\bf Safeguards}
    \item[] Question: Does the paper describe safeguards that have been put in place for responsible release of data or models that have a high risk for misuse (e.g., pretrained language models, image generators, or scraped datasets)?
    \item[] Answer: \answerNA{} 
    \item[] Justification: We believe that our work does not pose such risks.

\item {\bf Licenses for existing assets}
    \item[] Question: Are the creators or original owners of assets (e.g., code, data, models), used in the paper, properly credited and are the license and terms of use explicitly mentioned and properly respected?
    \item[] Answer: \answerYes{} 
    \item[] Justification: We adequately cited existing works.

\item {\bf New assets}
    \item[] Question: Are new assets introduced in the paper well documented and is the documentation provided alongside the assets?
    \item[] Answer: \answerYes{} 
    \item[] Justification: We provide detailed README instructions for our code in~\url{https://github.com/kswoo97/FGPAnom}.

\item {\bf Crowdsourcing and research with human subjects}
    \item[] Question: For crowdsourcing experiments and research with human subjects, does the paper include the full text of instructions given to participants and screenshots, if applicable, as well as details about compensation (if any)? 
    \item[] Answer: \answerNA{} 
    \item[] Justification: We do not contain any crowdsourcing and human-subjects-related experiments.

\item {\bf Institutional review board (IRB) approvals or equivalent for research with human subjects}
    \item[] Question: Does the paper describe potential risks incurred by study participants, whether such risks were disclosed to the subjects, and whether Institutional Review Board (IRB) approvals (or an equivalent approval/review based on the requirements of your country or institution) were obtained?
    \item[] Answer: \answerNA{} 
    \item[] Justification: We do not contain any crowdsourcing and human-subjects-related experiments.

\item {\bf Declaration of LLM usage}
    \item[] Question: Does the paper describe the usage of LLMs if it is an important, original, or non-standard component of the core methods in this research? Note that if the LLM is used only for writing, editing, or formatting purposes and does not impact the core methodology, scientific rigorousness, or originality of the research, declaration is not required.
    \item[] Answer: \answerNA{} 
    \item[] Justification: We only used LLMs to improve the writing.

\end{enumerate}

\label{sec:checklist}



\end{document}